\documentclass[10pt,journal,compsoc]{IEEEtran}
\newif\ifpeerreview

\peerreviewfalse

\usepackage[nocompress]{cite}
\usepackage{url}
\usepackage{amsmath,amssymb,amsfonts,graphicx}
\usepackage{siunitx}
\usepackage[euler]{textgreek}
\DeclareMathOperator*{\argmin}{arg\,min}

\usepackage{cite}
\usepackage{url}
\usepackage{lipsum}
\usepackage{amsmath}
\usepackage[normalem]{ulem}

\usepackage{array, makecell} 
\usepackage{blindtext}
\usepackage[utf8]{inputenc}
\usepackage[table]{xcolor}
\usepackage[flushleft]{threeparttable} 
\usepackage{subcaption}
\usepackage{soul}
\usepackage{tabularx,ragged2e,booktabs,caption}
\usepackage{adjustbox}
\usepackage[usestackEOL]{stackengine}
\usepackage{lipsum} % Only used to generate random text.
\usepackage[switch]{lineno}

\newcommand{\paperID}{54}

\title{Physics vs. Learned Priors: Rethinking Camera and Algorithm Design for Task-Specific Imaging}
\newcommand{\superscript}[1]{\ensuremath{^{\textrm{#1}}}}
\author{Tzofi Klinghoffer\superscript{*},
        Siddharth Somasundaram\superscript{*}, 
        Kushagra Tiwary\superscript{*}, Ramesh Raskar% <-this % stops a space
\IEEEcompsocitemizethanks{
\IEEEcompsocthanksitem \textbf{\superscript{*}} Equal contribution (alphabetical order)\protect
\IEEEcompsocthanksitem Each author is with Massachusetts Institute of Technology}% <-this % stops an unwanted space
}

\begin{document}

\IEEEtitleabstractindextext{%
\begin{abstract}
Cameras were originally designed using physics-based heuristics to capture aesthetic images. In recent years, there has been a transformation in camera design from being purely physics-driven to increasingly data-driven and task-specific. In this paper, we present a framework to understand the building blocks of this nascent field of \emph{end-to-end design} of camera hardware and algorithms. As part of this framework, we show how methods that exploit both physics and data have become prevalent in imaging and computer vision, underscoring a key trend that will continue to dominate the future of task-specific camera design. Finally, we share current barriers to progress in end-to-end design, and hypothesize how these barriers can be overcome.
\end{abstract}

}

% Make Title
\ifpeerreview
\linenumbers \linenumbersep 15pt\relax 
\author{Paper ID \paperID\IEEEcompsocitemizethanks{\IEEEcompsocthanksitem This paper is under review for ICCP 2022 and the PAMI special issue on computational photography. Do not distribute.}}
\markboth{Anonymous ICCP 2022 submission ID \paperID}%
{}
\fi
\maketitle
% Following line added to remove line numbers per ICCP guidance
\thispagestyle{empty}

% The first section title should be wrapped inside a \IEEEraisesectionheading as follows.
\IEEEraisesectionheading{
  \section{Introduction}\label{sec:introduction}
}

Early advances in imaging were inspired by human vision. For example, the Bayer filter was inspired by human retinal sensitivity to red, green, and blue light. Such insights from human vision have led to camera designs that have yielded remarkable digital photography capabilities and have widespread consumer applications.

The goals of \emph{imaging} have since shifted from photography to solving tasks such as 3D shape reconstruction, phase estimation, and material estimation. These tasks rely on information beyond what the human eye can directly measure, so it no longer makes sense to design imaging systems based on the eye. Much like the evolution of animal vision, camera design has evolved to adapt to the needs of the task and environment \cite{cronin2014visual}. By using known physics of light-matter interactions, physical cues such as polarization, interference, and spectrum are exploited to encode task-relevant information. Measurements of these cues can then be decoded into the scene parameter of interest by solving a model inversion problem. This idea of jointly exploiting physical cues and computation is the premise of the field of \emph{computational imaging}. 

Whereas imaging deals with capturing image representations of the world, \emph{computer vision} extracts meaningful information from these images for high-level tasks, such as classification, detection, and segmentation. The modern era of computer vision was ushered in by advances in sensors, computing, and algorithms. High-resolution sensors paved the way to megapixel resolution, computing systems provided the bandwidth needed to process high-dimensional data, and deep learning provided a framework to learn from large amounts of data. 

\begin{figure*}[t!]
  \centering
  \includegraphics[width=0.70\textwidth]{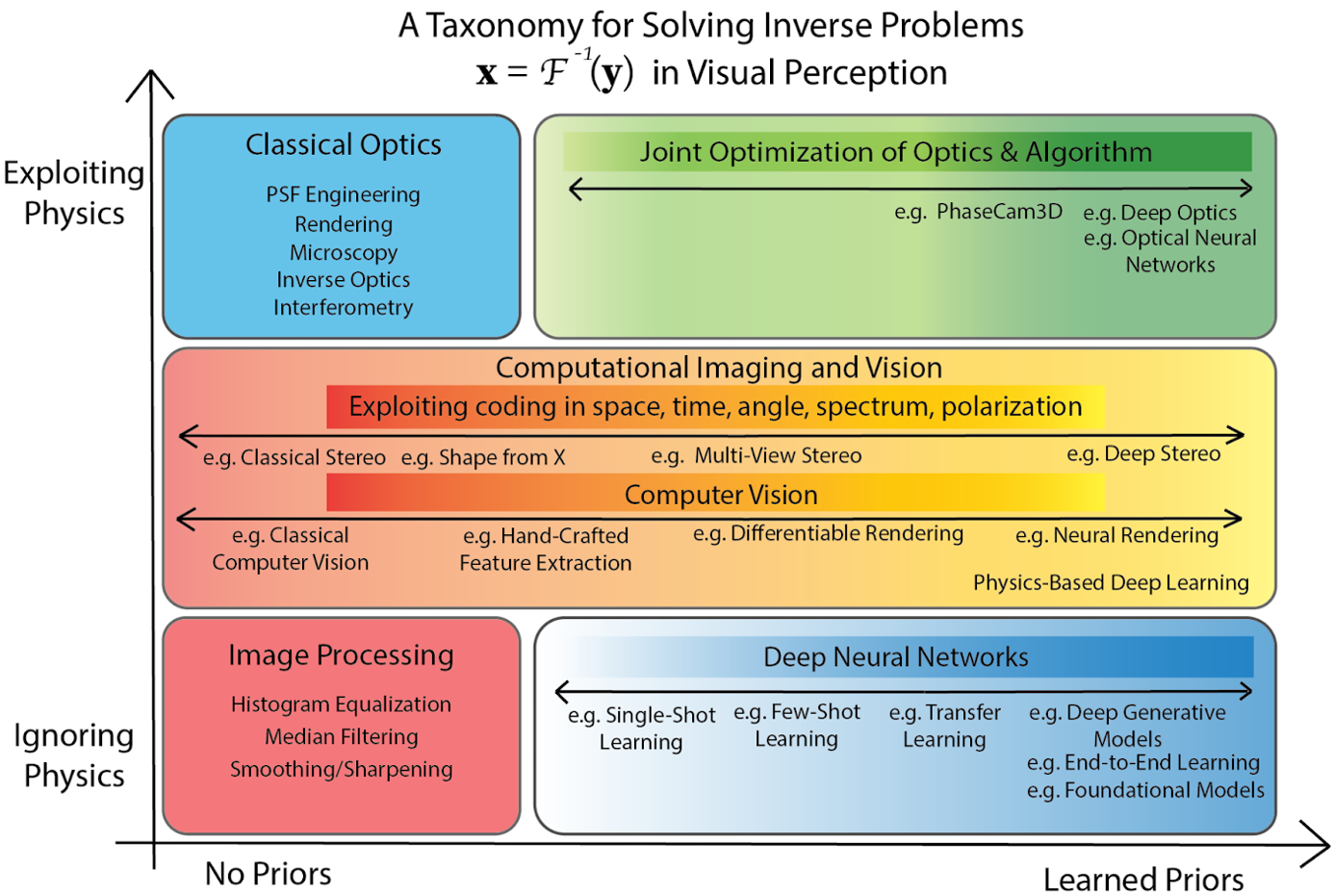}
  \caption{\textbf{Our framework maps inverse problems in visual perception by how they parametrize $\mathcal{F}$.} Deep learning focuses on learning priors through data-driven methods, whereas classical computer vision, optics, and computational imaging rely on physics. Each section of our paper corresponds to a field/method shown in this chart. We anticipate future imaging systems will use physics and data for \textit{joint optimization} (green box).
  }
  \label{fig:physics_data}
\end{figure*}

Together, these two steps, \textit{imaging} and \textit{vision}, form the visual perception pipeline. The imaging step can be interpreted as a physical encoder, transforming the 3D scene into a lower-dimensional representation (i.e. image), and the vision step as a digital decoder, using the representation to perform high-level tasks, as shown in Fig. \ref{fig:e2e}. While perception is more closely linked with vision, imaging has a downstream effect on perception that has historically been ignored. Recent works in \emph{end-to-end design} have brought together the design of the physical encoder (i.e. camera) and digital decoder (i.e. neural network) for task-specific perception. 

We envision design of future imaging systems to be done in an \emph{end-to-end} manner. Consider the problem of face recognition. A computational imaging approach would enable acquisition of a robust representation of the face (e.g. 3D shape and color from stereo), but inherently cannot perform face recognition. Vision approaches can learn from data to recognize faces, but will perform poorly in edge cases (e.g. low light). Intelligent and integrated design of both the imaging and vision steps can enable acquisition of robust image representations that lead to high-accuracy face recognition.

We discuss two major trends in this paper: 

\begin{enumerate}
    \item Visual perception is being solved in an \emph{end-to-end}, task-specific manner, with the imaging hardware jointly optimized with the vision algorithm.
    \item Imaging, while traditionally physics-based, has recently become increasingly data-driven to perform optimally on high-level vision tasks. Similarly, deep learning, while typically data-driven, has begun to incorporate physical models to drive advances in areas such as neural rendering \cite{sota_rend_tft}. 
\end{enumerate}

In connection to these trends, we discuss advances in five major fields related to visual perception (as shown in Fig. \ref{fig:physics_data}): image processing, classical optics, physics-based learning, computational imaging and vision, and joint optimization of optics and algorithm. Each of these topics is treated as a separate section in this paper, except for image processing and deep neural networks, for which we refer the reader to \cite{wiley2018computer} and \cite{lecun2015deep}, respectively. While other review perspectives highlight the use of optics for computing \cite{wetzstein2020inference}, we focus on the use of computing for optics. Task-specific cameras that are jointly designed with vision algorithms will continue to be a theme in the next generation of visual perception systems.

%The next generation of visual perception systems will be dominated by task-specific cameras that are jointly designed with vision algorithms. 
% In this paper, we provide a framework to understand the building blocks of modern end-to-end optimization. We then identify barriers to progress in end-to-end design, and predict exciting directions in perception that will arise out of a synergistic use of physics and data. 

\subsection{Contributions}

\begin{itemize}
    \item We present a review perspective on the building blocks of \emph{end-to-end design}, including a conceptual framework to connect and understand each block based on its use of physics and learned priors.
    \item We define five ingredients for designing end-to-end imaging systems, highlighted in Fig. \ref{fig:tools}. We also identify open challenges and hypothesize solutions along each axis.
\end{itemize}

\vspace{-2mm}

\section{Optical Camera Design}~\label{sec:analytical_design}
The most primitive camera is arguably the pinhole camera. A pinhole camera consists of a tiny aperture placed in front of the image plane to map a single scene point to a single image point, based on ray optics. In practice, diffraction effects and SNR limitations of a tiny aperture cause us to use lenses to focus light. More complex multi-lens systems enable extended depth of field and correction of optical aberrations. These imaging systems were derived from fundamental ideas in optics. In this section, we discuss mathematical- and simulation-based methods for designing optical imaging systems.

\subsection{Inverse Optics}
The initial goal of camera design was to efficiently map light coming from one point in a scene to one point on the image sensor using compound optics. Engineers design optical systems with a desired input-output pair by combining a sequence of optical components. They may leverage insights from first order optics using approximations such as thin lenses, paraxial approximations, and plane/spherical wave models. They then verify the design by using commercial raytracing software such as Zemax \cite{geary2002introduction} or Code V \cite{walker2008optical}. However, such design is often a process of trial and error.

\subsection{PSF Engineering}

\begin{figure}
  \centering
  \includegraphics[width=0.5\textwidth]{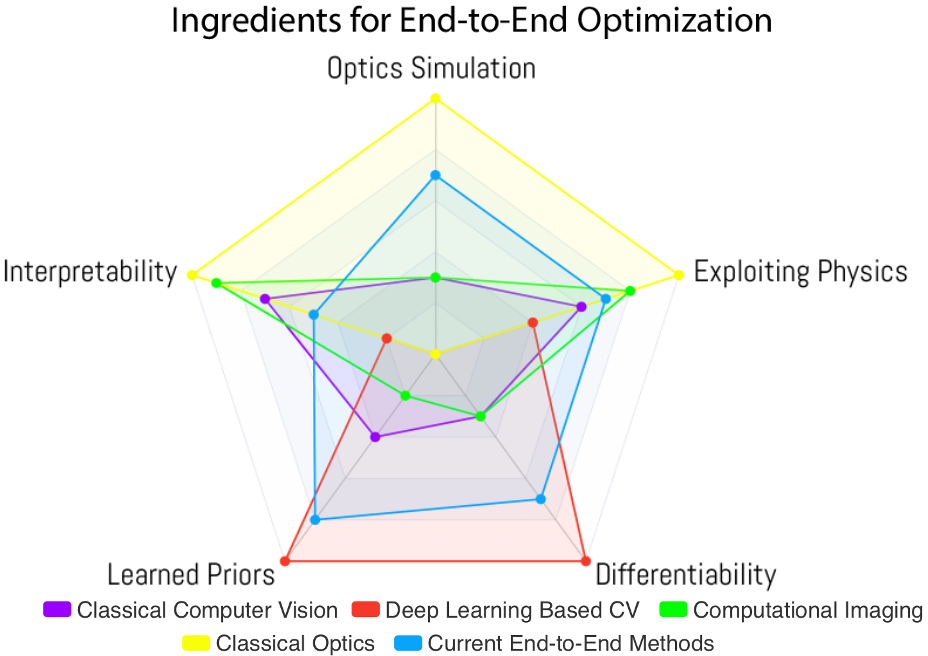}
  \caption{\textbf{Ingredients and open challenges for end-to-end methods.} We present five ingredients that we consider most important when designing end-to-end imaging systems, and evaluate how each field (computer vision, computational imaging, optics, current end-to-end methods) takes advantage of these ingredients. We anticipate future imaging systems will make more effective use of the five ingredients and span the entire radar chart. We discuss progress and challenges for each ingredient in Section \ref{sec:challenges}.}
  \label{fig:tools}
\end{figure}

The \textit{point spread function} (PSF) of an imaging system is its response to a point source of light, or the 2D spatial impulse response. A measured image $y$ can be expressed as 
\begin{equation}
    y = f(x * h + \eta),
\end{equation}

\begin{figure}
    \centering
    \includegraphics[width=0.4\textwidth]{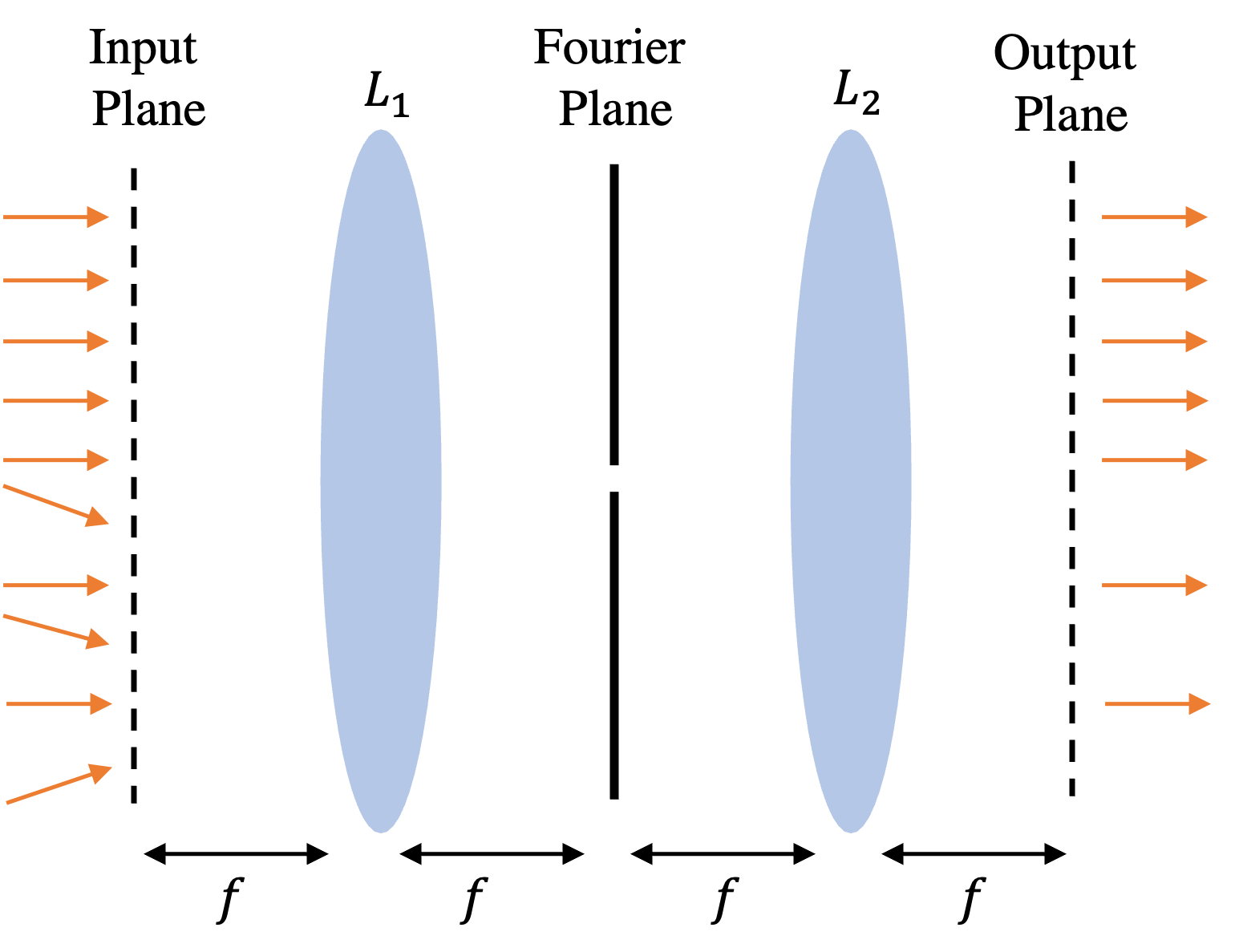}
    \caption{\textbf{4f system: a simple example of optical signal processing.} The mask at the Fourier plane is a pinhole here, which means the system behaves as a low-pass filter.}
    \label{fig:4f-system}
\end{figure}

where $x$ is the actual scene, $f(\cdot)$ is the camera response function, $*$ is the convolution operator, h is the PSF, and $\eta$ is read noise. $x$ can be estimated by solving the deconvolution problem in the Fourier domain. However, under certain conditions, the deconvolution problem is ill-posed and certain assumptions must be employed. Carefully designing the PSF (i.e. PSF engineering) can numerically guarantee more robust reconstruction of $x$. PSF engineering is the process of \textit{wavefront coding}, in which the electromagnetic field distribution at the aperture plane is modified in a known way to encode information about the scene. $h$ is physically constructed by refractive elements, phase masks, and amplitude masks. 

PSF engineering has applications in microscopy for single-particle tracking \cite{wulstein2017point, opatovski2021multiplexed, shechtman2016multicolour}. It has also been employed for extended depth of field \cite{dowski1995extended}. We will discuss approaches that enable optimization of the PSF. There are works that optimize the PSF using deep learning \cite{nehme2020deepstorm3d}, but we restrict our discussion to analytical methods in this section. 

\subsubsection{Fourier Domain Approaches}
One of the simplest examples of PSF engineering is the manipulation the optical signal in the 2D spatial frequency domain. This manipulation can be done using a 4f optical system, as shown in Fig. \ref{fig:4f-system}. A lens takes a 2D Fourier transform of the incident beam. At the Fourier plane, an amplitude (or phase) mask can be placed to manipulate the relative intensities (or phases) of different frequency components. Another lens takes the inverse Fourier transform of the beam before it is fed to the sensor. In this case, the mask at the Fourier plane determines the optical transfer function (OTF), which is the Fourier Transform of the PSF. The 4f system is frequently used for optical signal filtering. For a detailed reference on Fourier optics, the reader is directed to the introductory text by Goodman \cite{goodman1968introduction}.

\subsubsection{Information Theory Approaches}
The goal of imaging is to obtain some scene property, $x$, from a set of observations, $y$. Doing so requires us to design an "optically informative" PSF. Fisher information and statistical information theory give us a method of doing so. Fisher information gives us a way to quantify the information content of a PSF by measuring the sensitivity of $y$ to changes in $x$, given a PSF. The Fisher information matrix can be used to compute Cram\'er Rao lower bounds on the variance of the estimation. This method has been applied in microscopy for single particle tracking \cite{shechtman2014optimal}, fluorescence imaging beyond the diffraction limit \cite{pavani2009three}, and depth estimation \cite{greengard2005fisher}.

\subsubsection{Wigner Distribution and Ambiguity Function}
The \textit{space-bandwidth product theorem} states that the product of the moduli of a function and its Fourier transform must be greater than or equal to some constant. In ray optics, this theorem translates to the idea that there will exist a tradeoff between the accuracy in measuring the position and direction of a ray. The \textit{Wigner distribution} enables a visualization of the Fourier conjugate variables: complex electric field amplitude and spatial frequency ($k$ vector). The \textit{ambiguity function} is the Fourier dual of the Wigner distribution. For a more detailed explanation of the Wigner distribution, the interested reader is directed to \cite{alonso2011wigner, testorf2010phase}. Horstmeyer \emph{et al.} use the ambiguity function to design an optimal depth-dependent PSF \cite{horstmeyer2011partially}.

\section{Computational Imaging}

%%%%%%%%%%%%%%%%%%%%%%%%%%%%%%%%%%%%%%%%%%%%%%%
% Begin Table
%%%%%%%%%%%%%%%%%%%%%%%%%%%%%%%%%%%%%%%%%%%%%%%
\begin{table*}\label{tab:comp_img}

\setlength{\arrayrulewidth}{.3 mm}
\setlength{\tabcolsep}{5pt}
\renewcommand{\arraystretch}{3}
\newcolumntype{s}{p{1cm}}
\arrayrulecolor[RGB]{0,0,0}
\fontsize{8}{8}\selectfont
\begin{center}
\begin{threeparttable}
\begin{tabular}{ c|c|c|c|c|c| }
\cline{2-6}
 & \Centerstack{\cellcolor[RGB]{250,175,175}\textbf{Space}} &\cellcolor[RGB]{250,175,175}\Centerstack{\textbf{Time}} &\cellcolor[RGB]{250,175,175}\Centerstack{\textbf{Angle}}&\cellcolor[RGB]{250,175,175}\Centerstack{\textbf{Spectrum}}&\cellcolor[RGB]{250,175,175}\Centerstack{\textbf{Polarization}} \\

\hline
\multicolumn{1}{|c|}{\cellcolor[RGB]{175,250,175}\Centerstack{\textbf{Coding with} \\ \textbf{Illumination}}} &  \cellcolor[RGB]{250,250,175} \Centerstack{Direct-Global \\ Separation \cite{nayar2006fast} \\ Photometric Stereo \cite{woodham1980photometric}} & \cellcolor[RGB]{250,250,175} \Centerstack{Light Transport \\ \cite{tanaka2018time, martin1980picosecond} \\ Multipath \\ Interference \cite{kadambi2013coded} \\ Fluorescence \\ Imaging \cite{bhandari2015blind}} &  \cellcolor[RGB]{250,250,175} \Centerstack{Refraction \\ Measurement \cite{wetzstein2011hand}}  & \cellcolor[RGB]{250,250,175} \Centerstack{Multispectral \\ Imaging\cite{park2007multispectral} \\ Low-Light \\ \cellcolor[RGB]{250,250,175} Photography \cite{krishnan2009dark}} &  \cellcolor[RGB]{250,250,175} \Centerstack{Descattering \cite{schmitt1992use}} \\

\hline
\multicolumn{1}{|c|}{ \cellcolor[RGB]{175,250,175}\Centerstack{\textbf{Coding with} \\ \textbf{Optics}}} &  \cellcolor[RGB]{250,250,175} \Centerstack{HDR Imaging \cite{martel2020neural} \\ \cellcolor[RGB]{250,250,175} Single-Shot Depth \\ \cellcolor[RGB]{250,250,175} \cite{levin2007image, wu2019phasecam3d, yanny2020miniscope3d}} & \cellcolor[RGB]{250,250,175} \Centerstack{Time Stretch \\ Imaging \cite{mahjoubfar2017time} \\ Motion \\ Deblurring \cite{raskar2006coded}} & \cellcolor[RGB]{250,250,175} \Centerstack{Digital \\ Refocusing \cite{ng2005light} \\ Novel View \\ Estimation \cite{levoy2006light}} & \cellcolor[RGB]{250,250,175} \Centerstack{Multispectral \\ Imaging \\ \cite{themelis2008multispectral, monno2017adaptive, alvarez2016practical, baek2017compact, shogenji2004multispectral}} & \cellcolor[RGB]{250,250,175} \Centerstack{Shape Estimation \\ \cite{taamazyan2016shape, kadambi2015polarized} \\ Stokes Imaging \cite{zhao2010liquid} \\ Light Transport \\ Decomposition \cite{nayar1997separation, schechner1999polarization}} \\

\hline
\multicolumn{1}{|c|}{\cellcolor[RGB]{175,250,175}\Centerstack{\textbf{Coding with}\\\textbf{Sensors}}} & \cellcolor[RGB]{250,250,175} \Centerstack{Stereo Vision \cite{hartley2003multiple} \\ HDR Imaging \cite{bhandari2017unlimited} \\ Gradient Camera \cite{tumblin2005want}}  & \cellcolor[RGB]{250,250,175} \Centerstack{Depth Estimation (ToF) \\ NLOS Imaging \cite{velten2012recovering} \\ Imaging Through \\ Scattering \cite{satat2016all}} & \cellcolor[RGB]{250,250,175} \cellcolor[RGB]{250,250,175} \Centerstack{Wavefront \\ Sensing \cite{platt2001history}} & \cellcolor[RGB]{250,250,175} \Centerstack{Spatio-Spectral \\ Superresolution \cite{cao2011high} \\ Seeing Occluded \\ Objects \cite{adib2013see, maeda2019thermal}} & \cellcolor[RGB]{250,250,175} \Centerstack{3D Imaging \cite{kadambi2015polarized}} \\

% \hline
% \multicolumn{1}{|c|}{\Centerstack{\textbf{Passive Spatial}\\\textbf{Coherence}}} & None  & \Centerstack{Dual Phase Sagnac \\ Interferometer }& \cellcolor[RGB]{250,250,175} Medium &\cellcolor[RGB]{250,175,175} Sensitive  & \cellcolor[RGB]{250,175,175}\Centerstack{Scene Geometry}\\

% \hline
% \multicolumn{1}{|c|}{\Centerstack{\textbf{Active}\\\textbf{Coherence}}} & \Centerstack{Coherent Source} & \Centerstack{Traditional \\ Camera}& \cellcolor[RGB]{175,250,175} Low & \cellcolor[RGB]{250,175,175} Sensitive &\cellcolor[RGB]{250,175,175}\Centerstack{Scene Geometry}\\

% \hline
% \multicolumn{1}{|c|}{\Centerstack{\textbf{Passive}\\\textbf{Intensity}}} & \Centerstack{None} & \Centerstack{Traditional \\ Camera}& \cellcolor[RGB]{175,250,175} Low & \cellcolor[RGB]{250,175,175} Sensitive &\cellcolor[RGB]{250,175,175}\Centerstack{Scene Geometry \\ Occlusion Geometry}\\

% \hline
% \multicolumn{1}{|c|}{\makecell{\textbf{Active}\\\textbf{Intensity}}} & \Centerstack{Flashlight \\ Laser} & \Centerstack{Traditional\\Camera}& \cellcolor[RGB]{175,250,175} Low & \cellcolor[RGB]{250,175,175} Sensitive & \cellcolor[RGB]{250,175,175}\Centerstack{Scene Geometry}\\

\hline
\end{tabular}
\end{threeparttable}
\caption{\textbf{A classification of computational imaging problems based on how physics is encoded.} Note that some works can fall into multiple categories. This is not an exhaustive list, but is meant to illustrate some key examples.}
\end{center}
\vspace{-3mm}
\end{table*}

%%%%%%%%%%%%%%%%%%%%%%%%%%%%%%%%%%%%%%%%%%%%%%%
% End Table
%%%%%%%%%%%%%%%%%%%%%%%%%%%%%%%%%%%%%%%%%%%%%%%

Light is well described by the \emph{plenoptic dimensions}: space, time, angle, and spectrum. The goal of plenoptic imaging is to capture information about each of these dimensions, but in practice, most cameras capture only a subset of this information for a given application \cite{bergen1991plenoptic}. For example, the intensity $I(\mathbf{r})$ measured by a standard intensity camera can be expressed by the following integral

\begin{equation}
    I(\mathbf{r}) = \int I(\mathbf{r}, \mathbf{\Theta}, t, \lambda) \, d\mathbf{\Theta} \, dt \, d\lambda,
\end{equation}

where $\mathbf{r}=(x, y)$ is the spatial position, $\mathbf{\Theta}=(\theta, \phi)$ is the angle of incident light (i.e. light field), $t$ is the time of arrival of the light, and $\lambda$ is the wavelength. The plenoptic dimensions are commonly used in graphics because they enable a convenient representation of light in ray space. Additional wave properties, such as polarization, coherence, and phase, are also frequently used outside of graphics. 

Each dimension of light (space, time, angle, spectrum, polarization, phase, etc.) provides different information about the scene. For example, the time of flight (ToF) can be used to estimate depth, spectrum is strongly tied to material properties, polarization has fundamental relationships to shape and texture, and light field probing is tied to image focusing and depth of field. 

\emph{Computational imaging} is the co-design of imaging hardware and reconstruction algorithms for a vision application. The hardware is used to encode information about the scene in an image measurement. The algorithm is then designed to decode this scene information. We direct the interested reader to \cite{bhandari2022computational} and \cite{mait2018computational} for a more comprehensive treatment of computational imaging.

\begin{figure*}[t!]
  \centering
  \includegraphics[width=1.0\textwidth]{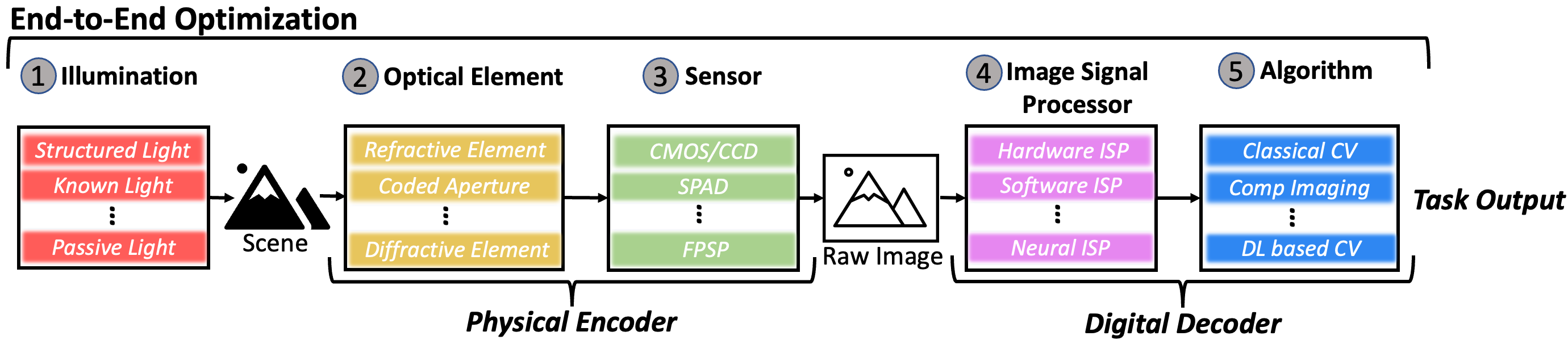}
  %\caption{\textbf{Modularity in end-to-end optimization over techniques spanning computer vision, computational imaging, and classical optics.} The components of imaging systems are designed and developed in silo. We formulate end-to-end optimization in a modular fashion and present it as a combination of known techniques that exist in computer vision, computational imaging, and classical optics that are jointly optimized to solve a downstream task. With this formulation, novel combinations of end-to-end systems can be created and optimized.}
  \caption{\textbf{Modular formulation for end-to-end optimization of the imaging and vision pipeline.} Components can be selected for each numbered module and then jointly optimized. Existing end-to-end methods optimize modules 2-4 with 5 (described in Section \ref{sec:AI_design}). The early parts of the image formation pipeline are implemented in hardware and encode the scene into RAW measurements (\emph{physical encoder}). Then, the \emph{digital decoder} produces the desired output, e.g. HDR image, classification, etc. With this proposed formulation, end-to-end systems can be created that use novel, non-obvious combinations of computer vision, computational imaging, and optics that are jointly optimized to solve downstream tasks.}
  
  %Numbered components can be optimized in existing end-to-end optimization methods and include the optical element, sensor, image signal processor (ISP), and neural network for downstream task. The early parts of the image formation pipeline are implemented in hardware and encode the scene into RAW pixel measurements, and thus can be considered a \emph{physical encoder}. The latter part of the pipeline decode the RAW pixel measurements into the desired output, i.e. HDR image, segmentation map, bounding boxes, classification, etc, and thus can be considered a \emph{digital decoder}. While the image signal processor (ISP) is shown as part of the digital decoder, it can be implemented as hardware or software, as described in Section \ref{sec:ISP_opt}.}
  \label{fig:e2e}
\end{figure*}

A general forward imaging model is described in Fig. \ref{fig:e2e}. A light source first illuminates the scene. The scene scatters light back to the imaging system, where the light is optically transformed by a configuration of optical elements. Then, light is measured by a sensor, which digitizes analog values into a digital image. Finally, the image is processed by an algorithm to provide perceptual information about the scene. From this, we see there are four degrees of freedom when designing computational imaging systems: illumination, optics, sensor, and algorithms. We discuss some examples of each degree of freedom in this section. We categorize each example in Table 1. The image signal processor (ISP) can be considered an additional degree of freedom, which we abstract into the algorithm in this section and discuss in section \ref{sec:ISP_opt}.

%can be considered an additional degree of freedom, but is abstracted into the algorithm in this section.

%\add{image signal processor (ISP)},
%, except ISP which is discussed in section \ref{sec:ISP_opt}.

\subsection{Coding with Illumination}
 \textit{Active illumination} is the process of illuminating the scene with a known and controlled pattern of light, as opposed to \textit{passive illumination} (i.e. ambient light). Active illumination enables more robust reconstruction of scene properties and, in some cases, enables faster image acquisition because light sources can be dynamically manipulated more easily than optical elements in front of the sensor. 
 
 Nayar \emph{et al.} use a high-frequency checkerboard pattern to decompose the light transport of a scene into direct and global components \cite{nayar2006fast}. Park \emph{et al.} acquire multispectral images by capturing multiple images with different multiplexed spectral sources  \cite{park2007multispectral}. Krishnan and Fergus use flash at near-infrared wavelengths for low-light photography and obtain reconstructions at visible wavelengths using spatio-spectral correlations \cite{krishnan2009dark}. Wetzstein \emph{et al.} measure the position and angle of light rays when probed with a light field to measure materials with varying refractive fields \cite{wetzstein2011hand}. Kadambi \emph{et al.} use custom temporal codes for robust depth measurements in the presence of multi-path interference \cite{kadambi2013coded}. Bhandari \emph{et al.} use time-coded illumination to estimate fluorescence lifetime with continuous wave ToF sensors \cite{bhandari2015blind}.
 
\emph{Photometric stereo (PS)} is a technique to obtain the surface normals of an object under controlled lighting \cite{woodham1980photometric}. In its most primitive form, PS illuminates a diffuse object sequentially with point sources from multiple lighting directions and the camera in a fixed position. The resultant shading profile can then be fit to a Lambertian model to intensity measurements. Subsequent works have since extended PS to non-diffuse materials and uncontrolled lighting \cite{basri2007photometric, alldrin2008photometric}. Tanaka \emph{et al.} use transient profiles of thermal light to extract diffuse reflectance of objects and perform photometric stereo, leveraging that the propagation speed of heat is resolvable at video frame rates \cite{tanaka2018time}. 

One way to image through \emph{scattering media} is by illuminating with modulated light. Light that returns to the sensor in the same modulated state is considered unscattered. For example, if light is linearly polarized and emitted at a scene immersed in a scattering media, the sensor would only measure light arriving at the linear polarization angle with strongest intensity, and reject all other light (i.e. polarization gating) \cite{schmitt1992use}. The same principle can also be applied with coherent light through scattering media, where only coherent light is measured (i.e. coherence gating). These principles are based on the idea that scattered light reduces degree of polarization and coherence. ToF imaging can also be used (i.e. time gating) since unscattered light (i.e. ballistic photons) has the shortest pathlength, meaning it arrives at the sensor first \cite{martin1980picosecond}.

\subsection{Coding with Optics}
Optical coding refers to optically transforming light arriving from the scene before it reaches the sensor. This transformation can be performed by optical components such as refractive elements (e.g. lenses), phase and amplitude masks, and color filter arrays.  

A classic example of optical coding is the \textit{Bayer filter}. By creating a mosaic of red, green, and blue pixels, an RGB image can be estimated by interpolation (i.e. demosaicking). Similarly, a multispectral image can be obtained by placing a multispectral filter array in front of the sensor, instead of a color filter array \cite{themelis2008multispectral, monno2017adaptive}. A multispectral image can also be obtained by placing diffraction gratings \cite{alvarez2016practical}, prisms \cite{baek2017compact}, or compound optics \cite{shogenji2004multispectral} in front of the sensor. In a similar spirit to the Bayer filter, full Stokes imaging can be performed by using a micropolarimeter, where micropolarizers at different linear polarization orientations are placed in front of each pixel \cite{zhao2010liquid}. Ng \emph{et al.} place a microlens array in front of a camera to capture a 4D light field, which enables digital refocusing of an image \cite{ng2005light}. It is also possible to generate novel views using a 4D light field \cite{levoy2006light}.

As discussed earlier, a well-chosen PSF can make the deconvolution problem well-posed. A coded exposure can be used to solve the problem of motion deblurring \cite{raskar2006coded} and for HDR imaging \cite{onzon2021neural}. A \textit{coded aperture} is an array of spatially varying opaque and transparent materials that modulate the phase and amplitude of incident light. Coded apertures were first deployed in X-ray imaging to provide a computational solution to forming X-rays, since X-rays cannot be focused with lenses like light \cite{caroli1987coded}. Since then, coded apertures have been applied to solve problems, such as single-shot depth \cite{levin2007image, wu2019phasecam3d, yanny2020miniscope3d}, high dynamic range (HDR) imaging \cite{martel2020neural, sun2020learning}, and 4D light field acquisition \cite{veeraraghavan2007dappled}. Some methods place a diffuser in front of a sensor for 3D reconstruction \cite{antipa2018diffusercam}, hyperspectral imaging \cite{monakhova2020spectral}, and microscopy \cite{liu2020fourier}. All these methods fall under \emph{lensless imaging}, in which scene information is encoded with an optical element rather than a bulky lens \cite{boominathan2022recent}.

The light transport of specular vs. diffuse light \cite{nayar1997separation} and real vs. virtual images (caused by transparent materials) \cite{schechner1999polarization} can be decomposed using polarization filters. Intuitively, this decomposition is based on the idea of \textit{cross-polarization}, where the undesirable reflectance component (e.g. specular reflection) is highly polarized and can be filtered out by placing a linear polarizer in front of the sensor at an orientation orthogonal to that polarization state. Polarization filters can also be used for measuring shape of objects, using known models of Fresnel reflection \cite{ba2020deep, kadambi2015polarized, taamazyan2016shape}. 

\subsection{Coding with Sensors}
The choice of sensor dictates what type of information can be captured in a scene. For example, a single photon avalanche detector (SPAD) or streak camera can be used to capture transients on the order of picoseconds. These time profiles can be used to capture light-in-flight videos, measure ToF, and analyze transient decays of fluorescent samples. Velten \emph{et al.} use ToF information and tomographic reconstruction for non-line-of-sight (NLOS) imaging. Other approaches to NLOS use thermal light emanating from occluded humans \cite{maeda2019thermal} and RF antennas to image through the occluder \cite{adib2013see}. Satat \emph{et al.} use ToF information to image through volumetric scattering by modeling scattering as a probabilistic spatio-temporal convolution \cite{satat2016all}. A Shack-Hartmann sensor can be used to measure the shape (i.e. phase distribution) of a wavefront \cite{platt2001history}.

Combining measurements from different sensing modalities, or \emph{sensor fusion}, has also shown great promise for different applications. Stereo vision uses two or more spatially offset RGB cameras that observe the same scene to obtain binocular cues, which are used for depth estimation \cite{hartley2003multiple}. Kadambi \emph{et al.} combine depth maps from a Kinect sensor with polarization images to obtain textures at micron scale \cite{kadambi2015polarized}. The idea is that depth maps are noisy, but polarization images alone can result in ambiguous geometries. Combining both modalities overcomes the individual weaknesses of each sensing modality. Similarly, Cao \emph{et al.} combine a high spatial resolution RGB image with a low spatial resolution multispectral image to obtain high spatial resolution multispectral images  \cite{cao2011high}. Ferstl \emph{et al.} use high spatial resolution intensity images to upsample low spatial resolution depth maps obtained from a Kinect sensor to estimate high resolution depth maps \cite{ferstl2013image}.

Manipulating backend electronics of the sensor also has interesting applications. Bhandari \emph{et al.} show that modulo sampling at the sensor level enables reconstruction of HDR images \cite{bhandari2017unlimited} \cite{bhandari2020unlimited}. \emph{Event cameras}, which measure changes in intensity rather than raw intensity, also enable HDR imaging \cite{rebecq2019high}. Tumblin \emph{et al.} manipulate the sensor to send the logarithm of the gradients of intensity to the analog-to-digital (ADC) converter (instead of intensity), which has applications for downstream tasks and HDR imaging \cite{tumblin2005want}. A coded two bucket (C2B) sensor is a sensor in which each pixel contains two "buckets", with each bucket measuring light with a programmable exposure. C2Bs have been used for compressive sensing \cite{martel2021computational} and 3D shape estimation \cite{wei2018coded}, among other applications. 

% \textbf{Image formation and perception pipeline used for end-to-end optimization.} Numbered components are optimized in existing end-to-end optimization methods and include the optical element, sensor, image signal processor (ISP), and neural network for downstream task. The early parts of the image formation pipeline are implemented in hardware and encode the scene into RAW pixel measurements, and thus can be considered a \emph{physical encoder}. The latter part of the pipeline decode the RAW pixel measurements into the desired output, i.e. HDR image, segmentation map, bounding boxes, classification, etc, and thus can be considered a \emph{digital decoder}. While the image signal processor (ISP) is shown as part of the digital decoder, it can be implemented as hardware or software, as described in Section \ref{sec:ISP_opt}.

\subsection{Forward Model Inversion}
An imaging system is typically modeled by a forward operator $\mathcal{F}(\cdot)$, such that the measured image, $y$, can be related to a scene parameter, $x$, by $y=\mathcal{F}(x)+\eta$. Solutions for $x$ can be found by solving an optimization problem of the form

\begin{equation}
    \hat{x} = \argmin_x \{||y - \mathcal{F}(x)||_2^2 +\alpha \phi(x)\},
\end{equation}

where $\phi(x)$ is a regularization term and $\alpha$ is the weight. If $\mathcal{F}$ is differentiable, we can solve for $x$ using gradient descent. This is the premise of \emph{differentiable optics} and \emph{differentiable rendering}. For a tutorial on classic optimization methods, we refer the reader to Boyd and Vandenberghe \cite{boyd2004convex}. Other methods use domain-specific algorithms or neural networks. We discuss these ideas in the following sections.

% \subsubsection{Coding in Space}
% \sid{flutter shutter, coded apertures (phase/amplitude masks), direct-global separation using high frequency illumination (Nayar)}

% \subsubsection{Coding in Time}
% \sid{ToF, fluorescence imaging, NLOS, LIDAR (Velodyne, Luminar, etc.)}

% \subsubsection{Coding in Angle}
% \sid{dappled photography, refocusing (Lytro)}

% \subsubsection{Coding in Polarization}
% \sid{SfP, diffuse/specular separation, imaging through scattering}

% \subsubsection{Coding in Spectrum}
% \sid{MSFA/compound optics, multiplexed spectral illumination, used for material classification}

% \subsubsection{Computational Microscopy}
% \sid{Fourier Ptychography, quantitative phase imaging, fluorescence microscopy}

\section{Physics-Based Computer Vision}
\label{sec:phys_based_vision}
Physics-based vision incorporates physical priors into otherwise data-driven deep learning methods, often by solving problems similar to forward model inversion, but through the use of a neural network. Deep learning approaches have largely dominated the computer vision community over the last decade, and involve the use of large amounts of data to iteratively learn an algorithm with backpropogation and gradient descent. In recent years, the incorporation of more \emph{physics} in deep learning vision models has unleashed new breakthroughs, such as in the area of neural rendering. Physics can be incorporated into deep learning vision models in several ways, as discussed in the rest of this section.
% In recent years, data-driven computer vision methods have become more reliant on physical models. 

\subsection{Physics-Based Learning}
Physics-based learning (PBL) is a group of methods that combine the use of physical priors with data-driven neural networks. We adopt the categorization of PBL methods, as proposed by \cite{ba2019blending}, into physical fusion, residual physics, embedded physics, and physical regularization. 

\emph{Physical fusion} models use the output of physical models as input to neural networks, which are then optimized for a downstream task, and have been employed in applications such as shape from polarization \cite{ba2020deep}. \emph{Residual physics}, on the other hand, uses the output of a physical model as ground truth to guide the training of a neural network. Applications of residual physics include training a neural network to create CT reconstructions from limited views, using filtered back projection (FBP) to generate ground truth as done by Jin \textit{et al.} \cite{jin2017deep}. \emph{Embedded physics} deals with methods in which the neural network is tasked with learning the parameters of a physical model, such as in algorithm unrolling, where each layer corresponds with an iteration of an iterative model \cite{monakhova2019learned}. Lastly, we explore \emph{physics regularization} in more depth in Section \ref{sec:phys_reg}, as it is used for a variety of vision tasks. 

Additional PBL methods have also been proposed that incorporate physics into deep learning in other ways, such as using simulated data from a physics model to pre-train a neural network for better weight initialization \cite{jia2021physics, willard2020integrating}.

%\sid{PhysicsNAS seems slightly misplaced, but I don't feel strongly about it. + 1 KT }
% such as physics neural architecture search (NAS) \cite{ba2019blending}, which includes physical forward models as part of the search space during NAS, and
% Physics regularization is a set of methods that adds an additional physics-based term to the loss function. \sid{"Often, physics-based loss terms are derived from estimating the parameters of a physical system." This feels like imprecise wording.} 

\subsubsection{Physics Regularization}
~\label{sec:phys_reg}
A loss function consists of a primary term and regularization term(s). Physics regularization uses known physics of the scene parameter $x$ and enforces this as an additional term in the loss function. In such a setup, a neural network is often used to predict physical parameters from input data, such as images. A loss can then be applied on the predicted parameters, or they can be used by the physical model to create additional output(s). For example, in the work of Che \textit{et al.} \cite{che2018inverse}, the physical scene parameters necessary to reconstruct an image with a renderer, such as shape, material, and illumination, are predicted and compared to known ground truth parameter values. In general, access to ground truth parameters of a physical model may not be available, motivating methods that predict parameters of a physical model and enforce their accuracy only through the final output, such as image reconstruction. These methods are commonly used for inverse and neural rendering, discussed further in Section \ref{sec:neural_rend_data}. Physics regularization has also been used for areas such as image dehazing by estimating parameters of the atmospheric scattering model \cite{Morales_2019_CVPR_Workshops}, deblurring \cite{chen2018reblur2deblur}, and depth prediction \cite{fei2019geo}, to name a few.

\subsection{Differentiable Rendering}
\label{sec:diff_rend}
%Rendering is the process of forming an image from the properties that define a scene, such as geometry, camera parameters, illumination model, material properties, etc, and thus falls in the \textbf{known physics} regime. Recent advances have made once non-differentiable renderers now differentiable, enabling ...

Rendering in graphics uses known models of the world, including scene geometry, illumination models, material properties, etc., to render the scene from the camera's perspective. The field of inverse graphics aims to recover geometry, reflectance, material properties and illumination from images of the scene. \emph{Differentiable rendering} stems from the field of inverse graphics, where the forward model is made differentiable, thus enabling the scene parameters to be recovered by computing gradients and performing gradient descent. OpenDR ~\cite{opendr} is one of the earliest implementations of this idea. Although it assumes a simple lighting model and cannot render complex effects such as inter-reflections, OpenDR paved the way for an entire class of differentiable renderers, such as SoftRas \cite{liu2019soft}, Mitsuba2 \cite{nimier2019mitsuba}, PyRedner \cite{redner}, among others \cite{Zhang:2020:PSDR} \cite{volrender}, which can handle ever more complex forward light simulations. While differentiable rendering can be used in isolation, the framework can also work alongside neural networks. The machine learning component typically predicts the scene parameters such as geometry \cite{ss_mesh_pred}, textures, or diffuse and specular reflections. More recently, neural networks have been used to encode the scene directly. We discuss this transition to using more data and machine learning in the following section.

\subsection{Neural Rendering}
\label{sec:neural_rend_data}
Neural rendering refers to the field that emerges from the use of neural networks to learn scene parameters, which can then be used generate photo-realistic imagery in a controllable way \cite{neural_rendering}. 2D neural rendering often relies on statistical approaches for image synthesis, such as deep generative models, in contrast to graphics frameworks that traditionally rely on light transport and the rendering equation to render images \cite{sota_rend_tft} \cite{10.1145/15886.15902}. Approaches that rely on deep generative models typically use generative adversarial networks (GANs) \cite{gan_review}, variational auto-encoders (VAEs) \cite{vae_review}, or diffusion models \cite{diffusion_models} to learn a distribution by training on the task of image-to-image translation. Therefore, these methods are data-driven and do not exploit physics. Recent 2D neural rendering techniques have merged physically inspired modules into data-driven methods by using techniques from computer graphics, such as illumination models \cite{Meka:2019}, textures \cite{10.1145/3306346.3323035} and multi-view constraints \cite{Lombardi:2018} \cite{PGZED19}, among others. Exploiting physics-based priors gives greater control over image synthesis over purely data-driven methods, by allowing light, geometry, and camera to be disentangled. For a comprehensive review for 2D neural rendering, we refer the reader to \cite{sota_rend_tft}.

The field of 3D neural rendering has also seen tremendous growth primarily due to the use of differentiable renderers discussed in Section \ref{sec:diff_rend}. 3D neural rendering aims to learn and render a 3D neural scene representation from real-world imagery and relies on the image formation model by leveraging techniques from computer graphics. State-of-the-art techniques parameterize the scene with a neural network using a differentiable renderer for novel-view synthesis. Typically, volumetric rendering is used for the differentiable rendering component as its continuous representation has been shown to work well with gradient descent \cite{volrender} \cite{nerf}, although there are a plethora of techniques that use different scene representations and rendering techniques \cite{neural_rendering, sota_rend_tft}. This physics and machine learning framework has been highly effective and subsequent works have added additional physics-based priors such as reflectance models \cite{nerv}, \cite{boss2021nerd}, normal estimation \cite{kuang2021neroic}, and shadow models \cite{https://doi.org/10.48550/arxiv.2203.15946} to enable better novel-view synthesis and 3D reconstruction. Moreover, these physics-based priors are now also used to train on classical computer vision tasks, such as object classification and segmentation, and show improved performance over purely data-driven techniques \cite{https://doi.org/10.48550/arxiv.2104.08418} \cite{https://doi.org/10.48550/arxiv.2104.01148}. Neural de-rendering has also been used for unsupervised representation learning, and has led to improved downstream accuracy over purely data-driven methods \cite{klinghoffer22physically}.

%Other work attempts to estimate the parameters of graphics renderers as a proxy task for learning meaningful visual representations with utility across downstream tasks \cite{klinghoffer22physically}.
%One class of methods seeks to use physical constraints to improve performance of computer vision tasks. Rather than directly predict labels to target tasks, neural networks are trained to predict intermediary physical parameters of a scene. These physical parameters constrain the neural network's predictions to conform with known physics. For example, work in image dehazing estimates parameters of the atmospheric scattering model to reconstruct a dehazed image \cite{Morales_2019_CVPR_Workshops}. Particle-based representations are also used to model interactions between objects of different materials by modeling 3D points that make up objects in a scene with rigity labels\cite{li2020visual}. In addition to modeling rigid and fluid materials, geometry-based constraints are also used to ensure dynamics of rigid body systems are consistent through time (isola's paper).

\section{Joint Optimization of Optics \& Algorithm}~\label{sec:AI_design}
\label{sec:joint_opt_baby}

Imaging systems have typically been designed independently of the downstream perception task that they seek to solve. \emph{Joint optimization}, also known as \emph{end-to-end optimization} or \emph{deep optics}, seeks to address this by jointly optimizing optics and image processing together for either low-level imaging or high-level vision tasks. Deep optics has been applied to low-level problems, such as color imaging and demosaicking \cite{chakrabarti2016learning}, extended depth of field and superresolution imaging \cite{Sitzmann:2018:EndToEndCam}, high dynamic range (HDR) imaging \cite{metzler2020deep} \cite{sun2020learning}, and depth estimation \cite{chang2019deep,haim2018depth,he2018learning}, and high level problems, such as classification \cite{chang2018hybrid} and object detection \cite{robidoux2021end} \cite{onzon2021neural}, \cite{del2020learned}. Deep optics has also been used in time of flight imaging \cite{marco2017deeptof,su2018deep,nehme2019dense}, computational microscopy \cite{hershko2019multicolor,horstmeyer2017convolutional,kellman2019data}, and imaging through scattering \cite{turpin2018light}. Existing work primarily focuses on optimizing the parameters of the optical element, sensor, and image signal processor (ISP), each of which can be considered a core block of the imaging pipeline and can be filled with different components in a plug-and-play manner. Optimization of each block is described in the following sections.

%, each described in more detail below.

\subsection{Optical Element Optimization}

The first component of the imaging pipeline that is considered for end-to-end optimization is the optical element. Advances in optimization and autodifferentiation have enabled the optics parameters of imaging systems to be optimized for downstream imaging and perception tasks. Sitzmann \emph{et al.} \cite{Sitzmann:2018:EndToEndCam} propose the joint optimization of the parameters of an imaging system's optical element and reconstruction algorithm by defining a fully differentiable wave optics image formation model. They experiment with optimizing a Zernike parameterization and Fourier coefficient parameterization of a lens using gradient-based optimization in simulation. Once optimized in simulation, the optical element is manufactured and results are validated in the real world. While the work of Sitzmann \emph{et al.} \cite{Sitzmann:2018:EndToEndCam} is applied to learning optimal parameters for the tasks of achromatic extended depth of field and super-resolution imaging, the proposed framework is task-agnostic and can be extended to optimize for other downstream tasks. \cite{metzler2020deep} extend this method to optimize optics for high dynamic range (HDR) imaging, while \cite{chang2019deep} extend this method to optimize optics for depth estimation. Ikoma \textit{et al.} \cite{ikoma2021depth} optimize the aperture design for monocular depth estimation. Tseng \textit{et al.} \cite{tseng2021differentiable} optimize the compound optics, i.e. parameters of multiple lenses, along with the ISP and downstream neural network, in a fully end-to-end pipeline. After optimal lens parameters are learned in simulation, the lens can be manufactured and used in production systems.

Rather than optimize an optical element, other methods replace the optical elements altogether and instead optimize a phase mask, such as in the work of Wu \emph{et al.} \cite{wu2019phasecam3d}. The phase mask is manufactured with photolithography, and is used to change the phase of light passing through it. Wu \emph{et al.} \cite{wu2019phasecam3d} introduce a framework for jointly learning the optimal phase mask and reconstruction algorithm for depth estimation.

\subsection{Sensor Optimization}

Recent advances in image sensor technology have resulted in the capability for per-pixel sensing and processing, unleashing new opportunities for sensor optimization \cite{martel2021computational}. Martel \textit{et al.} \cite{martel2020neural} propose a method to learn spatially varying pixel exposures for the tasks of HDR and high-speed imaging. Rather than use expensive spatial light modulators (SLMs) to implement optical coding, exposure times are directly learned and then programmed onto a focal-plane sensor–processor (FPSP), which can run both sensing and processing for each pixel. In their work, a per-pixel shutter function is parameterized by a neural network and learned with stochastic gradient descent in simulation before being programmed onto the sensor. The shutter function encodes the exposure time of each pixel. Nguyen \textit{et al.} \cite{nguyen2022learning} adopt a similar method for motion deblurring through spatially varying pixel exposures. Li \textit{et al.} \cite{li2020end} propose a different way to optimize the sensor for compressive sensing (CS). In their work, a CS mask is directly learned by a neural network to encode the scene with fewer bits and then programmed on a coded two-bucket (C2B) camera \cite{wei2018coded}.

%Rather than use costly spatial light modulators (SLMs) to implement optical coding, Martel \textit{et al.} \cite{martel2020neural} learn  spatially varying pixel exposures for the tasks of HDR and high-speed imaging, running inference aboard a focal-plane sensor–processor. In their work, a per-pixel shutter function is parameterized by a neural network and learned with stochastic gradient descent. The shutter function encodes the exposure time of each pixel.
%optimize spatially varying pixel exposures for the tasks of HDR and high-speed imaging. Rather than use a fixed exposure time for each pixel as is traditionally done, modern digital sensors may use optical coding to spatially vary pixel exposures, but usually require expensive  spatial light modulators (SLMs) for optical coding. Martel \textit{et al.} \cite{martel2020neural} overcome the need for using an SLM by directly learning the spatially varying pixel exposures jointly with the downstream task. They use focal-plane sensor–processors, which enable per-pixel sensing and processing, 
%elements of the sensor used to form the RAW image. These methods typically optimize the compressive sensing mask or spatially varying pixel exposures.

\subsection{Image Signal Processor (ISP) Optimization}
~\label{sec:ISP_opt}

Image signal processors (ISPs) produce color images from RAW pixel measurements from the sensor through a pipeline of demosaicking, denoising, deblurring, tone mapping, etc. \cite{ramanath2005color}. ISPs can either be implemented on hardware, as traditionally done, or in software, which, while easier to optimize, is often slower. Hardware ISPs contain hundreds of categorical and continuous parameters that can be tuned based on application. However, because the inner workings of hardware ISPs are usually not revealed to users, imaging experts are needed to manually configure the parameters, which can be time-intensive. As a result, methods that can automatically identify the optimal values of ISP parameters for a given downstream task, such as object detection, are of interest. While grid search can be used to optimize the parameters of a hardware ISP when there is a small number of parameters, the search space grows exponentially as more parameters are added. More sophisticated optimizations, such as the artificial bee colony algorithm have also been used for hardware ISP optimization as done by Nishimura \textit{et al.} \cite{nishimura2018automatic}, but were demonstrated on only 3-4 continuous parameters. Because the ISP is implemented in hardware and is non-differentiable, gradient-based approaches cannot be directly applied. Tseng \textit{et al.} \cite{tseng2019hyperparameter} observe that ISPs can instead be made differentiable by modeling them with neural networks. Thus, since the inner workings of hardware ISPs are hidden to protect trade secrets, the hardware ISP can be treated as a black box and optimized using a differentiable proxy function, i.e. neural network. Tseng \textit{et al.} use hardware in the training loop to acquire RAW pixel measurements of projected images and randomly sample hardware ISP parameters. The RAW pixel measurements and sampled hardware ISP parameters are used to train a neural network to generate the same output as the hardware ISP. Once trained, the parameters of the hardware ISP can be directly optimized with gradient descent for a given downstream task.

Robidoux \textit{et al.} \cite{robidoux2021end} propose an alternative method for hardware ISP optimization in the framework of end-to-end optimization of multi-exposure high dynamic range (HDR) camera systems. In their work, optimization alternates between 0\textsuperscript{th}-order evolutionary search over sensor and ISP parameters and 1\textsuperscript{st}-order gradient descent on the neural network weights, where the neural network is trained for a perception task, such as automotive object detection, using images captured by the camera system. An evolutionary search algorithm similar to CMA-ES (Covariance Matrix Adaptation Evolution Strategy) \cite{jastrebski2006improving} is used to optimize the hardware ISP parameters. To optimize the model, raw data is collected with 254 different ISP parameters and simulated SNR drop artifacts are added.

A variety of other methods attempt to replace the image processing steps done by the ISP with a single optimized or learned operation \cite{heide2014flexisp,jiang2017learning,chen2018learning,diamond2021dirty}. Diamond \textit{et al.} \cite{diamond2021dirty} propose Anscombe networks as neural ISPs that transform RAW pixel measurements to an output image, and jointly train the neural ISP with a network for the downstream task. They find that directly learning the downstream task from RAW pixel measurements leads to worse performance than when either a hardware or neural ISP is used, and that neural ISPs are especially helpful in low-light conditions where traditional ISPs produce worse output.

\subsection{Optical Neural Networks}
Optical neural networks (ONNs) are physical implementations of neural networks using optical components. Because operations are done with photons rather than electrons, ONNs may bring major increases in bandwidth. Early research found that fully connected layers can be cheaply implemented with optics components \cite{farhat1985optical}, followed by more recent advances in implementing convolutional layers in optics \cite{chang2018hybrid}. However, because optics operations are typically linear, implementing nonlinearity, an essential component to the success of deep neural networks, with optics is an active area of research. For more information on ONNs, we refer the reader to \cite{sui2020review} and \cite{wetzstein2020inference}.

% \section{E2E for Sensing and Robotics}~\label{sec:e2e_robotics}
% \subsection{Joint Optimization for End-to-End Robotics}
% \subsubsection{Differentiable Rendering}
% % \subsection{Joint Optimization of Sensing and Robotisc}
% \subsection{Physics-Based AI in Other Fields }

\begin{figure}
    \centering
    \includegraphics[width=0.5\textwidth]{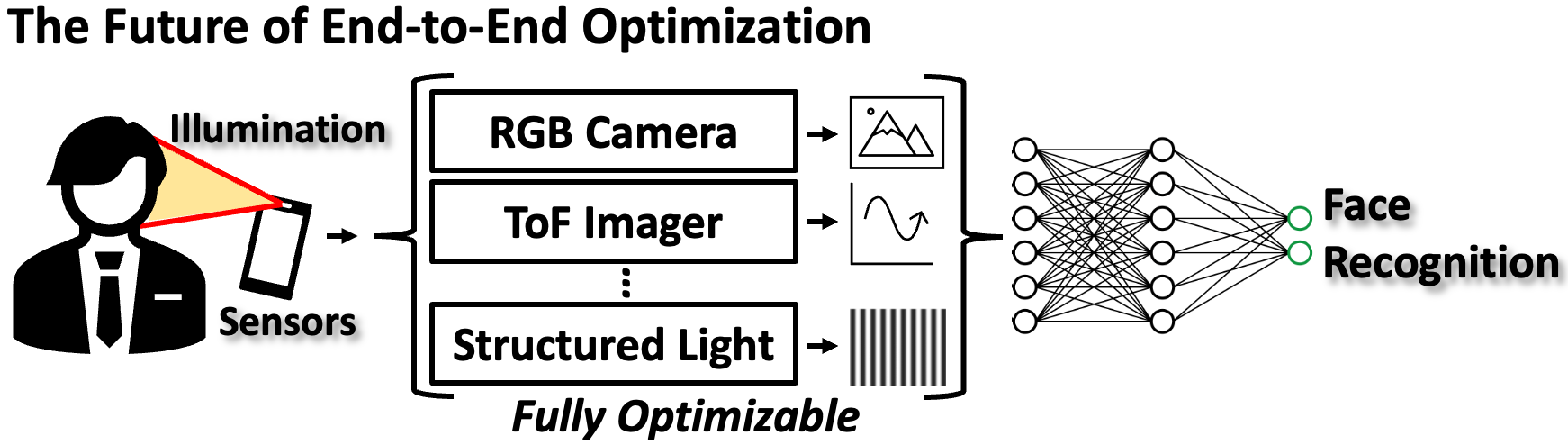}
    \caption{\textbf{Future of Camera System Design.} Progress in joint optimization of camera hardware and software has enabled task-specific cameras. In the future, this trend will continue and extend to fully end-to-end optimizable imaging \emph{systems} that choose which modalities to use to achieve a task.}
    \label{fig:future_e2e}
\end{figure}

\section{Ingredients for Joint Optimization}
We present five ingredients that we anticipate will drive progress in end-to-end methods and rank relevant fields for each ingredient in Fig. \ref{fig:tools}. These ingredients are \textbf{optical simulators}, \textbf{differentiability}, \textbf{interpretability}, \textbf{learned priors}, and \textbf{exploitation of physics}. While there has been significant progress in advancing each tool, we identify open challenges and present some insights into overcoming them.

% \vspace{5mm}
\subsection{Progress and Open Challenges}
~\label{sec:challenges}
%In this paper, we present a toolkit to conceptualize methods for camera and algorithm design that exploit physics and data. In Figure \ref{fig:tools}, we present five tools that are highly relevant to joint optimization of both optics and algorithms. Furthermore, in Section \ref{sec:joint_opt_baby} we present state-of-the-art algorithms that rank differently in those 5 axes. Here we discuss the most important open challenges and barriers to success in this area.

\noindent
\textbf{Optical simulators}. Optical simulators like Zemax and techniques like PSF engineering with deep learning have enabled data- and task-driven optical design. Yet, optical simulation remains a core challenge for end-to-end methods. Simulating complex optics of a camera is difficult and often proprietary. One solution to this challenge is to create open source simulators that can drive research and lower the barrier of entry for new researchers. Data-driven simulators, such as \cite{zhang2020image}, can help in overcoming this challenge.

% Lack of access to such simulators can slow the research and development process. \textbf{Solution:} By sharing simulators developed in research and building up a shared library of tools, 
    
\noindent
\textbf{Differentiability}: Although differentiability is not imperative for optimization, recent progress in computer vision, as discussed in Section \ref{sec:diff_rend}, has shown that differentiable systems converge faster and are more optimal. Advances in autodifferentiation have enabled progress in end-to-end design. Nevertheless, the problem of \textit{how} much differentiability is needed and how to make complex simulators differentiable is an open question for these systems. 
% and how much is needed to optimize end-to-end imagers
% We discussed End-to-End differentiability, or methods that are able to backpropogate from task output to the optics component, in Section \ref{sec:joint_opt_baby}.

\noindent
\textbf{Interpretability}: As task-specific imaging systems that learn over data and physics become ubiquitous, interpretability of the systems decrease. Many end-to-end methods analyze the learned PSFs of the system or perform analysis on intermediate outputs. Nevertheless, while the physical encoder remains more interpretable than pure deep learning systems, interpretability will remain a challenge as more learned priors are used. We can begin to overcome this by leveraging known physics and inputs, and comparing outputs from learned models to those from physical models.

% raw images under low-light, among other edge cases

%physics and testing the system's behaviour to known inputs/outputs, and perturbing inputs based on physics of the underlying system. 

\noindent
\textbf{Learned Priors}: Ideally, the end-to-end imaging system will continually learn and self-optimize to generalize to new environments that it may not have encountered in simulated training data. The key challenge is the sim-to-real gap and labeling real-world data on-the-fly. Existing methods use static or simulated scenes to avoid expensive relabeling of large datasets. Therefore, we anticipate the need for multi-step optimization, where training first happens in simulation and is then evaluated on real-world data, using the evaluation score to inform continued training in simulation. Progress in continual and lifelong learning can be leveraged to create cameras that adapt and learn from their environment \cite{parisi2019continual}.

\noindent
\textbf{Exploiting Physics}: Physics can be exploited in hardware and algorithms. End-to-end frameworks exploit physics at the hardware level, but do not fully exploit physics at the algorithm level. We anticipate future work will use differentiable, physics-based forward models that encode properties, such as reflectance and geometry, for better convergence and performance.

\section{Discussion}
Camera design has typically been rooted in the physics of light transport, but there has been progress recently towards combining physics-based design with \emph{data-driven} design. We claim that both are essential tools for task-specific camera design. Simultaneously, modern computer vision, which typically relies on data-driven deep learning, has begun to incorporate physics into deep learning methods. We propose a framework that classifies state-of-art methods based on their use of physics and data/learned priors. We also identify five key ingredients for end-to-end design, and show how challenges in these five areas can be addressed. In the future, systems may intelligently choose both which types of imagers to use and how to configure them, as shown in Fig. \ref{fig:future_e2e}. We envision a modular end-to-end pipeline to drive non-intuitive camera design that employs principles from computational imaging, physics-based computer vision, and optics. 
% \sid{"For example, an optimized framework that uses neural rendering and leverages the advantages of light field imaging for downstream tasks might yield better novel-view synthesis." If its just one sentence tacked on at the end, don't think this adds any value.}

%\sid{"Both imaging and vision are moving towards the joint use of physics and data for optimal downstream performance." This is repeating previous two sentences"}

\bibliographystyle{IEEEtran}
\bibliography{references}

\begin{thebibliography}{100}
\providecommand{\url}[1]{#1}
\csname url@samestyle\endcsname
\providecommand{\newblock}{\relax}
\providecommand{\bibinfo}[2]{#2}
\providecommand{\BIBentrySTDinterwordspacing}{\spaceskip=0pt\relax}
\providecommand{\BIBentryALTinterwordstretchfactor}{4}
\providecommand{\BIBentryALTinterwordspacing}{\spaceskip=\fontdimen2\font plus
\BIBentryALTinterwordstretchfactor\fontdimen3\font minus
  \fontdimen4\font\relax}
\providecommand{\BIBforeignlanguage}[2]{{%
\expandafter\ifx\csname l@#1\endcsname\relax
\typeout{** WARNING: IEEEtran.bst: No hyphenation pattern has been}%
\typeout{** loaded for the language `#1'. Using the pattern for}%
\typeout{** the default language instead.}%
\else
\language=\csname l@#1\endcsname
\fi
#2}}
\providecommand{\BIBdecl}{\relax}
\BIBdecl

\bibitem{cronin2014visual}
T.~W. Cronin, S.~Johnsen, N.~J. Marshall, and E.~J. Warrant, ``Visual
  ecology,'' in \emph{Visual Ecology}.\hskip 1em plus 0.5em minus 0.4em\relax
  Princeton University Press, 2014.

\bibitem{sota_rend_tft}
\BIBentryALTinterwordspacing
A.~Tewari, O.~Fried, J.~Thies, V.~Sitzmann, S.~Lombardi, K.~Sunkavalli,
  R.~Martin-Brualla, T.~Simon, J.~Saragih, M.~Nießner, R.~Pandey, S.~Fanello,
  G.~Wetzstein, J.-Y. Zhu, C.~Theobalt, M.~Agrawala, E.~Shechtman, D.~B.
  Goldman, and M.~Zollhöfer, ``State of the art on neural rendering,'' 2020.
  [Online]. Available: \url{https://arxiv.org/abs/2004.03805}
\BIBentrySTDinterwordspacing

\bibitem{wiley2018computer}
V.~Wiley and T.~Lucas, ``Computer vision and image processing: a paper
  review,'' \emph{International Journal of Artificial Intelligence Research},
  vol.~2, no.~1, pp. 29--36, 2018.

\bibitem{lecun2015deep}
Y.~LeCun, Y.~Bengio, and G.~Hinton, ``Deep learning,'' \emph{nature}, vol. 521,
  no. 7553, pp. 436--444, 2015.

\bibitem{wetzstein2020inference}
G.~Wetzstein, A.~Ozcan, S.~Gigan, S.~Fan, D.~Englund, M.~Solja{\v{c}}i{\'c},
  C.~Denz, D.~A. Miller, and D.~Psaltis, ``Inference in artificial intelligence
  with deep optics and photonics,'' \emph{Nature}, vol. 588, no. 7836, pp.
  39--47, 2020.

\bibitem{geary2002introduction}
J.~M. Geary, \emph{Introduction to lens design: with practical ZEMAX
  examples}.\hskip 1em plus 0.5em minus 0.4em\relax Willmann-Bell Richmond, VA,
  USA:, 2002.

\bibitem{walker2008optical}
B.~H. Walker, \emph{Optical engineering fundamentals}.\hskip 1em plus 0.5em
  minus 0.4em\relax Spie Press Bellingham, 2008, vol.~82.

\bibitem{wulstein2017point}
D.~M. Wulstein and R.~McGorty, ``Point-spread function engineering enhances
  digital fourier microscopy,'' \emph{Optics letters}, vol.~42, no.~22, pp.
  4603--4606, 2017.

\bibitem{opatovski2021multiplexed}
N.~Opatovski, Y.~Shalev~Ezra, L.~E. Weiss, B.~Ferdman, R.~Orange-Kedem, and
  Y.~Shechtman, ``Multiplexed psf engineering for three-dimensional multicolor
  particle tracking,'' \emph{Nano Letters}, vol.~21, no.~13, pp. 5888--5895,
  2021.

\bibitem{shechtman2016multicolour}
Y.~Shechtman, L.~E. Weiss, A.~S. Backer, M.~Y. Lee, and W.~Moerner,
  ``Multicolour localization microscopy by point-spread-function engineering,''
  \emph{Nature photonics}, vol.~10, no.~9, pp. 590--594, 2016.

\bibitem{dowski1995extended}
E.~R. Dowski and W.~T. Cathey, ``Extended depth of field through wave-front
  coding,'' \emph{Applied optics}, vol.~34, no.~11, pp. 1859--1866, 1995.

\bibitem{nehme2020deepstorm3d}
E.~Nehme, D.~Freedman, R.~Gordon, B.~Ferdman, L.~E. Weiss, O.~Alalouf, T.~Naor,
  R.~Orange, T.~Michaeli, and Y.~Shechtman, ``Deepstorm3d: dense 3d
  localization microscopy and psf design by deep learning,'' \emph{Nature
  methods}, vol.~17, no.~7, pp. 734--740, 2020.

\bibitem{goodman1968introduction}
J.~W. Goodman, \emph{Introduction to Fourier Optics. Goodman}.\hskip 1em plus
  0.5em minus 0.4em\relax McGraw-Hill, 1968.

\bibitem{shechtman2014optimal}
Y.~Shechtman, S.~J. Sahl, A.~S. Backer, and W.~E. Moerner, ``Optimal point
  spread function design for 3d imaging,'' \emph{Physical review letters}, vol.
  113, no.~13, p. 133902, 2014.

\bibitem{pavani2009three}
S.~R.~P. Pavani, M.~A. Thompson, J.~S. Biteen, S.~J. Lord, N.~Liu, R.~J. Twieg,
  R.~Piestun, and W.~E. Moerner, ``Three-dimensional, single-molecule
  fluorescence imaging beyond the diffraction limit by using a double-helix
  point spread function,'' \emph{Proceedings of the National Academy of
  Sciences}, vol. 106, no.~9, pp. 2995--2999, 2009.

\bibitem{greengard2005fisher}
A.~Greengard and R.~Piestun, ``Fisher information of 3d rotating point spread
  functions,'' in \emph{Computational Optical Sensing and Imaging}.\hskip 1em
  plus 0.5em minus 0.4em\relax Optical Society of America, 2005, p. CMA1.

\bibitem{alonso2011wigner}
M.~A. Alonso, ``Wigner functions in optics: describing beams as ray bundles and
  pulses as particle ensembles,'' \emph{Advances in Optics and Photonics},
  vol.~3, no.~4, pp. 272--365, 2011.

\bibitem{testorf2010phase}
M.~Testorf, B.~Hennelly, and J.~Ojeda-Casta{\~n}eda, \emph{Phase-space optics:
  fundamentals and applications}.\hskip 1em plus 0.5em minus 0.4em\relax
  McGraw-Hill Education, 2010.

\bibitem{horstmeyer2011partially}
R.~Horstmeyer, S.~B. Oh, O.~Gupta, and R.~Raskar, ``Partially coherent
  ambiguity functions for depth-variant point spread function design,'' 2011.

\bibitem{nayar2006fast}
S.~K. Nayar, G.~Krishnan, M.~D. Grossberg, and R.~Raskar, ``Fast separation of
  direct and global components of a scene using high frequency illumination,''
  in \emph{ACM SIGGRAPH 2006 Papers}, 2006, pp. 935--944.

\bibitem{woodham1980photometric}
R.~J. Woodham, ``Photometric method for determining surface orientation from
  multiple images,'' \emph{Optical engineering}, vol.~19, no.~1, pp. 139--144,
  1980.

\bibitem{tanaka2018time}
K.~Tanaka, N.~Ikeya, T.~Takatani, H.~Kubo, T.~Funatomi, and Y.~Mukaigawa,
  ``Time-resolved light transport decomposition for thermal photometric
  stereo,'' in \emph{Proceedings of the IEEE Conference on Computer Vision and
  Pattern Recognition}, 2018, pp. 4804--4813.

\bibitem{martin1980picosecond}
J.~Martin, Y.~Lecarpentier, A.~Antonetti, and G.~Grillon, ``Picosecond laser
  stereometry light scattering measurements on biological material,''
  \emph{Medical and Biological Engineering and Computing}, vol.~18, no.~2, pp.
  250--252, 1980.

\bibitem{kadambi2013coded}
A.~Kadambi, R.~Whyte, A.~Bhandari, L.~Streeter, C.~Barsi, A.~Dorrington, and
  R.~Raskar, ``Coded time of flight cameras: sparse deconvolution to address
  multipath interference and recover time profiles,'' \emph{ACM Transactions on
  Graphics (ToG)}, vol.~32, no.~6, pp. 1--10, 2013.

\bibitem{bhandari2015blind}
A.~Bhandari, C.~Barsi, and R.~Raskar, ``Blind and reference-free fluorescence
  lifetime estimation via consumer time-of-flight sensors,'' \emph{Optica},
  vol.~2, no.~11, pp. 965--973, 2015.

\bibitem{wetzstein2011hand}
G.~Wetzstein, R.~Raskar, and W.~Heidrich, ``Hand-held schlieren photography
  with light field probes,'' in \emph{2011 IEEE International Conference on
  Computational Photography (ICCP)}.\hskip 1em plus 0.5em minus 0.4em\relax
  IEEE, 2011, pp. 1--8.

\bibitem{park2007multispectral}
J.-I. Park, M.-H. Lee, M.~D. Grossberg, and S.~K. Nayar, ``Multispectral
  imaging using multiplexed illumination,'' in \emph{2007 IEEE 11th
  International Conference on Computer Vision}.\hskip 1em plus 0.5em minus
  0.4em\relax IEEE, 2007, pp. 1--8.

\bibitem{krishnan2009dark}
D.~Krishnan and R.~Fergus, ``Dark flash photography,'' \emph{ACM Trans.
  Graph.}, vol.~28, no.~3, p.~96, 2009.

\bibitem{schmitt1992use}
J.~Schmitt, A.~Gandjbakhche, and R.~Bonner, ``Use of polarized light to
  discriminate short-path photons in a multiply scattering medium,''
  \emph{Applied optics}, vol.~31, no.~30, pp. 6535--6546, 1992.

\bibitem{martel2020neural}
J.~N. Martel, L.~K. Mueller, S.~J. Carey, P.~Dudek, and G.~Wetzstein, ``Neural
  sensors: Learning pixel exposures for hdr imaging and video compressive
  sensing with programmable sensors,'' \emph{IEEE Transactions on Pattern
  Analysis and Machine Intelligence}, vol.~42, no.~7, pp. 1642--1653, 2020.

\bibitem{levin2007image}
A.~Levin, R.~Fergus, F.~Durand, and W.~T. Freeman, ``Image and depth from a
  conventional camera with a coded aperture,'' \emph{ACM transactions on
  graphics (TOG)}, vol.~26, no.~3, pp. 70--es, 2007.

\bibitem{wu2019phasecam3d}
Y.~Wu, V.~Boominathan, H.~Chen, A.~Sankaranarayanan, and A.~Veeraraghavan,
  ``Phasecam3d—learning phase masks for passive single view depth
  estimation,'' in \emph{2019 IEEE International Conference on Computational
  Photography (ICCP)}.\hskip 1em plus 0.5em minus 0.4em\relax IEEE, 2019, pp.
  1--12.

\bibitem{yanny2020miniscope3d}
K.~Yanny, N.~Antipa, W.~Liberti, S.~Dehaeck, K.~Monakhova, F.~L. Liu, K.~Shen,
  R.~Ng, and L.~Waller, ``Miniscope3d: optimized single-shot miniature 3d
  fluorescence microscopy,'' \emph{Light: Science \& Applications}, vol.~9,
  no.~1, pp. 1--13, 2020.

\bibitem{mahjoubfar2017time}
A.~Mahjoubfar, D.~V. Churkin, S.~Barland, N.~Broderick, S.~K. Turitsyn, and
  B.~Jalali, ``Time stretch and its applications,'' \emph{Nature Photonics},
  vol.~11, no.~6, pp. 341--351, 2017.

\bibitem{raskar2006coded}
R.~Raskar, A.~Agrawal, and J.~Tumblin, ``Coded exposure photography: motion
  deblurring using fluttered shutter,'' in \emph{ACM SIGGRAPH 2006 Papers},
  2006, pp. 795--804.

\bibitem{ng2005light}
R.~Ng, M.~Levoy, M.~Br{\'e}dif, G.~Duval, M.~Horowitz, and P.~Hanrahan, ``Light
  field photography with a hand-held plenoptic camera,'' Ph.D. dissertation,
  Stanford University, 2005.

\bibitem{levoy2006light}
M.~Levoy, ``Light fields and computational imaging,'' \emph{Computer}, vol.~39,
  no.~8, pp. 46--55, 2006.

\bibitem{themelis2008multispectral}
G.~Themelis, J.~S. Yoo, and V.~Ntziachristos, ``Multispectral imaging using
  multiple-bandpass filters,'' \emph{Optics letters}, vol.~33, no.~9, pp.
  1023--1025, 2008.

\bibitem{monno2017adaptive}
Y.~Monno, D.~Kiku, M.~Tanaka, and M.~Okutomi, ``Adaptive residual interpolation
  for color and multispectral image demosaicking,'' \emph{Sensors}, vol.~17,
  no.~12, p. 2787, 2017.

\bibitem{alvarez2016practical}
S.~Alvarez-Cortes, T.~Kunkel, and B.~Masia, ``Practical low-cost recovery of
  spectral power distributions,'' in \emph{Computer Graphics Forum}, vol.~35,
  no.~1.\hskip 1em plus 0.5em minus 0.4em\relax Wiley Online Library, 2016, pp.
  166--178.

\bibitem{baek2017compact}
S.-H. Baek, I.~Kim, D.~Gutierrez, and M.~H. Kim, ``Compact single-shot
  hyperspectral imaging using a prism,'' \emph{ACM Transactions on Graphics
  (TOG)}, vol.~36, no.~6, pp. 1--12, 2017.

\bibitem{shogenji2004multispectral}
R.~Shogenji, Y.~Kitamura, K.~Yamada, S.~Miyatake, and J.~Tanida,
  ``Multispectral imaging using compact compound optics,'' \emph{Optics
  Express}, vol.~12, no.~8, pp. 1643--1655, 2004.

\bibitem{taamazyan2016shape}
V.~Taamazyan, A.~Kadambi, and R.~Raskar, ``Shape from mixed polarization,''
  \emph{arXiv preprint arXiv:1605.02066}, 2016.

\bibitem{kadambi2015polarized}
A.~Kadambi, V.~Taamazyan, B.~Shi, and R.~Raskar, ``Polarized 3d: High-quality
  depth sensing with polarization cues,'' in \emph{Proceedings of the IEEE
  International Conference on Computer Vision}, 2015, pp. 3370--3378.

\bibitem{zhao2010liquid}
X.~Zhao, A.~Bermak, F.~Boussaid, and V.~G. Chigrinov, ``Liquid-crystal
  micropolarimeter array for full stokes polarization imaging in visible
  spectrum,'' \emph{Optics express}, vol.~18, no.~17, pp. 17\,776--17\,787,
  2010.

\bibitem{nayar1997separation}
S.~K. Nayar, X.-S. Fang, and T.~Boult, ``Separation of reflection components
  using color and polarization,'' \emph{International Journal of Computer
  Vision}, vol.~21, no.~3, pp. 163--186, 1997.

\bibitem{schechner1999polarization}
Y.~Y. Schechner, J.~Shamir, and N.~Kiryati, ``Polarization-based decorrelation
  of transparent layers: The inclination angle of an invisible surface,'' in
  \emph{Proceedings of the seventh IEEE international conference on computer
  vision}, vol.~2.\hskip 1em plus 0.5em minus 0.4em\relax IEEE, 1999, pp.
  814--819.

\bibitem{hartley2003multiple}
R.~Hartley and A.~Zisserman, \emph{Multiple view geometry in computer
  vision}.\hskip 1em plus 0.5em minus 0.4em\relax Cambridge university press,
  2003.

\bibitem{bhandari2017unlimited}
A.~Bhandari, F.~Krahmer, and R.~Raskar, ``On unlimited sampling,'' in
  \emph{2017 International Conference on Sampling Theory and Applications
  (SampTA)}.\hskip 1em plus 0.5em minus 0.4em\relax IEEE, 2017, pp. 31--35.

\bibitem{tumblin2005want}
J.~Tumblin, A.~Agrawal, and R.~Raskar, ``Why i want a gradient camera,'' in
  \emph{2005 IEEE Computer Society Conference on Computer Vision and Pattern
  Recognition (CVPR'05)}, vol.~1.\hskip 1em plus 0.5em minus 0.4em\relax IEEE,
  2005, pp. 103--110.

\bibitem{velten2012recovering}
A.~Velten, T.~Willwacher, O.~Gupta, A.~Veeraraghavan, M.~G. Bawendi, and
  R.~Raskar, ``Recovering three-dimensional shape around a corner using
  ultrafast time-of-flight imaging,'' \emph{Nature communications}, vol.~3,
  no.~1, pp. 1--8, 2012.

\bibitem{satat2016all}
G.~Satat, B.~Heshmat, D.~Raviv, and R.~Raskar, ``All photons imaging through
  volumetric scattering,'' \emph{Scientific reports}, vol.~6, no.~1, pp. 1--8,
  2016.

\bibitem{platt2001history}
B.~C. Platt and R.~Shack, ``History and principles of shack-hartmann wavefront
  sensing,'' pp. S573--S577, 2001.

\bibitem{cao2011high}
X.~Cao, X.~Tong, Q.~Dai, and S.~Lin, ``High resolution multispectral video
  capture with a hybrid camera system,'' in \emph{CVPR 2011}.\hskip 1em plus
  0.5em minus 0.4em\relax IEEE, 2011, pp. 297--304.

\bibitem{adib2013see}
F.~Adib and D.~Katabi, ``See through walls with wifi!'' in \emph{Proceedings of
  the ACM SIGCOMM 2013 conference on SIGCOMM}, 2013, pp. 75--86.

\bibitem{maeda2019thermal}
T.~Maeda, Y.~Wang, R.~Raskar, and A.~Kadambi, ``Thermal non-line-of-sight
  imaging,'' in \emph{2019 IEEE International Conference on Computational
  Photography (ICCP)}.\hskip 1em plus 0.5em minus 0.4em\relax IEEE, 2019, pp.
  1--11.

\bibitem{bergen1991plenoptic}
J.~R. Bergen and E.~H. Adelson, ``The plenoptic function and the elements of
  early vision,'' \emph{Computational models of visual processing}, vol.~1,
  p.~8, 1991.

\bibitem{bhandari2022computational}
A.~Bhandari, A.~Kadambi, and R.~Raskar, \emph{Computational Imaging}, 2022.

\bibitem{mait2018computational}
J.~N. Mait, G.~W. Euliss, and R.~A. Athale, ``Computational imaging,''
  \emph{Advances in Optics and Photonics}, vol.~10, no.~2, pp. 409--483, 2018.

\bibitem{basri2007photometric}
R.~Basri, D.~Jacobs, and I.~Kemelmacher, ``Photometric stereo with general,
  unknown lighting,'' \emph{International Journal of computer vision}, vol.~72,
  no.~3, pp. 239--257, 2007.

\bibitem{alldrin2008photometric}
N.~Alldrin, T.~Zickler, and D.~Kriegman, ``Photometric stereo with
  non-parametric and spatially-varying reflectance,'' in \emph{2008 IEEE
  Conference on Computer Vision and Pattern Recognition}.\hskip 1em plus 0.5em
  minus 0.4em\relax IEEE, 2008, pp. 1--8.

\bibitem{onzon2021neural}
E.~Onzon, F.~Mannan, and F.~Heide, ``Neural auto-exposure for high-dynamic
  range object detection,'' in \emph{Proceedings of the IEEE/CVF Conference on
  Computer Vision and Pattern Recognition}, 2021, pp. 7710--7720.

\bibitem{caroli1987coded}
E.~Caroli, J.~Stephen, G.~Di~Cocco, L.~Natalucci, and A.~Spizzichino, ``Coded
  aperture imaging in x-and gamma-ray astronomy,'' \emph{Space Science
  Reviews}, vol.~45, no.~3, pp. 349--403, 1987.

\bibitem{sun2020learning}
Q.~Sun, E.~Tseng, Q.~Fu, W.~Heidrich, and F.~Heide, ``Learning rank-1
  diffractive optics for single-shot high dynamic range imaging,'' in
  \emph{Proceedings of the IEEE/CVF conference on computer vision and pattern
  recognition}, 2020, pp. 1386--1396.

\bibitem{veeraraghavan2007dappled}
A.~Veeraraghavan, R.~Raskar, A.~Agrawal, A.~Mohan, and J.~Tumblin, ``Dappled
  photography: Mask enhanced cameras for heterodyned light fields and coded
  aperture refocusing,'' \emph{ACM Trans. Graph.}, vol.~26, no.~3, p.~69, 2007.

\bibitem{antipa2018diffusercam}
N.~Antipa, G.~Kuo, R.~Heckel, B.~Mildenhall, E.~Bostan, R.~Ng, and L.~Waller,
  ``Diffusercam: lensless single-exposure 3d imaging,'' \emph{Optica}, vol.~5,
  no.~1, pp. 1--9, 2018.

\bibitem{monakhova2020spectral}
K.~Monakhova, K.~Yanny, N.~Aggarwal, and L.~Waller, ``Spectral diffusercam:
  Lensless snapshot hyperspectral imaging with a spectral filter array,''
  \emph{Optica}, vol.~7, no.~10, pp. 1298--1307, 2020.

\bibitem{liu2020fourier}
F.~L. Liu, G.~Kuo, N.~Antipa, K.~Yanny, and L.~Waller, ``Fourier diffuserscope:
  single-shot 3d fourier light field microscopy with a diffuser,'' \emph{Optics
  Express}, vol.~28, no.~20, pp. 28\,969--28\,986, 2020.

\bibitem{boominathan2022recent}
V.~Boominathan, J.~T. Robinson, L.~Waller, and A.~Veeraraghavan, ``Recent
  advances in lensless imaging,'' \emph{Optica}, vol.~9, no.~1, pp. 1--16,
  2022.

\bibitem{ba2020deep}
Y.~Ba, A.~Gilbert, F.~Wang, J.~Yang, R.~Chen, Y.~Wang, L.~Yan, B.~Shi, and
  A.~Kadambi, ``Deep shape from polarization,'' in \emph{European Conference on
  Computer Vision}.\hskip 1em plus 0.5em minus 0.4em\relax Springer, 2020, pp.
  554--571.

\bibitem{ferstl2013image}
D.~Ferstl, C.~Reinbacher, R.~Ranftl, M.~R{\"u}ther, and H.~Bischof, ``Image
  guided depth upsampling using anisotropic total generalized variation,'' in
  \emph{Proceedings of the IEEE international conference on computer vision},
  2013, pp. 993--1000.

\bibitem{bhandari2020unlimited}
A.~Bhandari, F.~Krahmer, and R.~Raskar, ``On unlimited sampling and
  reconstruction,'' \emph{IEEE Transactions on Signal Processing}, vol.~69, pp.
  3827--3839, 2020.

\bibitem{rebecq2019high}
H.~Rebecq, R.~Ranftl, V.~Koltun, and D.~Scaramuzza, ``High speed and high
  dynamic range video with an event camera,'' \emph{IEEE transactions on
  pattern analysis and machine intelligence}, vol.~43, no.~6, pp. 1964--1980,
  2019.

\bibitem{martel2021computational}
J.~Martel and G.~Wetzstein, ``Computational imaging with vision sensors
  embedding in-pixel processing,'' in \emph{2021 IEEE International Electron
  Devices Meeting (IEDM)}.\hskip 1em plus 0.5em minus 0.4em\relax IEEE, 2021,
  pp. 35--3.

\bibitem{wei2018coded}
M.~Wei, N.~Sarhangnejad, Z.~Xia, N.~Gusev, N.~Katic, R.~Genov, and K.~N.
  Kutulakos, ``Coded two-bucket cameras for computer vision,'' in
  \emph{Proceedings of the European Conference on Computer Vision (ECCV)},
  2018, pp. 54--71.

\bibitem{boyd2004convex}
S.~Boyd, S.~P. Boyd, and L.~Vandenberghe, \emph{Convex optimization}.\hskip 1em
  plus 0.5em minus 0.4em\relax Cambridge university press, 2004.

\bibitem{ba2019blending}
Y.~Ba, G.~Zhao, and A.~Kadambi, ``Blending diverse physical priors with neural
  networks,'' \emph{arXiv preprint arXiv:1910.00201}, 2019.

\bibitem{jin2017deep}
K.~H. Jin, M.~T. McCann, E.~Froustey, and M.~Unser, ``Deep convolutional neural
  network for inverse problems in imaging,'' \emph{IEEE Transactions on Image
  Processing}, vol.~26, no.~9, pp. 4509--4522, 2017.

\bibitem{monakhova2019learned}
K.~Monakhova, J.~Yurtsever, G.~Kuo, N.~Antipa, K.~Yanny, and L.~Waller,
  ``Learned reconstructions for practical mask-based lensless imaging,''
  \emph{Optics express}, vol.~27, no.~20, pp. 28\,075--28\,090, 2019.

\bibitem{jia2021physics}
X.~Jia, J.~Willard, A.~Karpatne, J.~S. Read, J.~A. Zwart, M.~Steinbach, and
  V.~Kumar, ``Physics-guided machine learning for scientific discovery: An
  application in simulating lake temperature profiles,'' \emph{ACM/IMS
  Transactions on Data Science}, vol.~2, no.~3, pp. 1--26, 2021.

\bibitem{willard2020integrating}
J.~Willard, X.~Jia, S.~Xu, M.~Steinbach, and V.~Kumar, ``Integrating
  physics-based modeling with machine learning: A survey,'' \emph{arXiv
  preprint arXiv:2003.04919}, vol.~1, no.~1, pp. 1--34, 2020.

\bibitem{che2018inverse}
C.~Che, F.~Luan, S.~Zhao, K.~Bala, and I.~Gkioulekas, ``Inverse transport
  networks,'' \emph{arXiv preprint arXiv:1809.10820}, 2018.

\bibitem{Morales_2019_CVPR_Workshops}
P.~Morales, T.~Klinghoffer, and S.~Jae~Lee, ``Feature forwarding for efficient
  single image dehazing,'' in \emph{Proceedings of the IEEE/CVF Conference on
  Computer Vision and Pattern Recognition (CVPR) Workshops}, June 2019.

\bibitem{chen2018reblur2deblur}
H.~Chen, J.~Gu, O.~Gallo, M.-Y. Liu, A.~Veeraraghavan, and J.~Kautz,
  ``Reblur2deblur: Deblurring videos via self-supervised learning,'' in
  \emph{2018 IEEE International Conference on Computational Photography
  (ICCP)}.\hskip 1em plus 0.5em minus 0.4em\relax IEEE, 2018, pp. 1--9.

\bibitem{fei2019geo}
X.~Fei, A.~Wong, and S.~Soatto, ``Geo-supervised visual depth prediction,''
  \emph{IEEE Robotics and Automation Letters}, vol.~4, no.~2, pp. 1661--1668,
  2019.

\bibitem{opendr}
M.~M. Loper and M.~J. Black, ``Opendr: An approximate differentiable
  renderer,'' in \emph{European Conference on Computer Vision}.\hskip 1em plus
  0.5em minus 0.4em\relax Springer, 2014, pp. 154--169.

\bibitem{liu2019soft}
S.~Liu, T.~Li, W.~Chen, and H.~Li, ``Soft rasterizer: A differentiable renderer
  for image-based 3d reasoning,'' in \emph{Proceedings of the IEEE/CVF
  International Conference on Computer Vision}, 2019, pp. 7708--7717.

\bibitem{nimier2019mitsuba}
M.~Nimier-David, D.~Vicini, T.~Zeltner, and W.~Jakob, ``Mitsuba 2: A
  retargetable forward and inverse renderer,'' \emph{ACM Transactions on
  Graphics (TOG)}, vol.~38, no.~6, pp. 1--17, 2019.

\bibitem{redner}
T.-M. Li, M.~Aittala, F.~Durand, and J.~Lehtinen, ``Differentiable monte carlo
  ray tracing through edge sampling,'' \emph{ACM Trans. Graph. (Proc. SIGGRAPH
  Asia)}, vol.~37, no.~6, pp. 222:1--222:11, 2018.

\bibitem{Zhang:2020:PSDR}
C.~Zhang, B.~Miller, K.~Yan, I.~Gkioulekas, and S.~Zhao, ``Path-space
  differentiable rendering,'' \emph{ACM Trans. Graph.}, vol.~39, no.~4, pp.
  143:1--143:19, 2020.

\bibitem{volrender}
J.~Weiss and N.~Navab, ``Deep direct volume rendering: Learning visual feature
  mappings from exemplary images,'' \emph{arXiv preprint arXiv:2106.05429},
  2021.

\bibitem{ss_mesh_pred}
\BIBentryALTinterwordspacing
Y.~Ye, S.~Tulsiani, and A.~Gupta, ``Shelf-supervised mesh prediction in the
  wild,'' 2021. [Online]. Available: \url{https://arxiv.org/abs/2102.06195}
\BIBentrySTDinterwordspacing

\bibitem{neural_rendering}
\BIBentryALTinterwordspacing
A.~Tewari, J.~Thies, B.~Mildenhall, P.~Srinivasan, E.~Tretschk, Y.~Wang,
  C.~Lassner, V.~Sitzmann, R.~Martin-Brualla, S.~Lombardi, T.~Simon,
  C.~Theobalt, M.~Niessner, J.~T. Barron, G.~Wetzstein, M.~Zollhoefer, and
  V.~Golyanik, ``Advances in neural rendering,'' 2021. [Online]. Available:
  \url{https://arxiv.org/abs/2111.05849}
\BIBentrySTDinterwordspacing

\bibitem{10.1145/15886.15902}
\BIBentryALTinterwordspacing
J.~T. Kajiya, ``The rendering equation,'' \emph{SIGGRAPH Comput. Graph.},
  vol.~20, no.~4, p. 143–150, aug 1986. [Online]. Available:
  \url{https://doi.org/10.1145/15886.15902}
\BIBentrySTDinterwordspacing

\bibitem{gan_review}
L.~Wang, W.~Chen, W.~Yang, F.~Bi, and F.~R. Yu, ``A state-of-the-art review on
  image synthesis with generative adversarial networks,'' \emph{IEEE Access},
  vol.~8, pp. 63\,514--63\,537, 2020.

\bibitem{vae_review}
\BIBentryALTinterwordspacing
H.~Huang, Z.~Li, R.~He, Z.~Sun, and T.~Tan, ``Introvae: Introspective
  variational autoencoders for photographic image synthesis,'' 2018. [Online].
  Available: \url{https://arxiv.org/abs/1807.06358}
\BIBentrySTDinterwordspacing

\bibitem{diffusion_models}
\BIBentryALTinterwordspacing
P.~Dhariwal and A.~Nichol, ``Diffusion models beat gans on image synthesis,''
  2021. [Online]. Available: \url{https://arxiv.org/abs/2105.05233}
\BIBentrySTDinterwordspacing

\bibitem{Meka:2019}
\BIBentryALTinterwordspacing
A.~Meka, C.~Haene, R.~Pandey, M.~Zollhoefer, S.~Fanello, G.~Fyffe, A.~Kowdle,
  X.~Yu, J.~Busch, J.~Dourgarian, P.~Denny, S.~Bouaziz, P.~Lincoln, M.~Whalen,
  G.~Harvey, J.~Taylor, S.~Izadi, A.~Tagliasacchi, P.~Debevec, C.~Theobalt,
  J.~Valentin, and C.~Rhemann, ``Deep reflectance fields - high-quality facial
  reflectance field inference from color gradient illumination,'' vol.~38,
  no.~4, July 2019. [Online]. Available:
  \url{http://gvv.mpi-inf.mpg.de/projects/DeepReflectanceFields/}
\BIBentrySTDinterwordspacing

\bibitem{10.1145/3306346.3323035}
\BIBentryALTinterwordspacing
J.~Thies, M.~Zollh\"{o}fer, and M.~Nie\ss{}ner, ``Deferred neural rendering:
  Image synthesis using neural textures,'' \emph{ACM Trans. Graph.}, vol.~38,
  no.~4, jul 2019. [Online]. Available:
  \url{https://doi.org/10.1145/3306346.3323035}
\BIBentrySTDinterwordspacing

\bibitem{Lombardi:2018}
S.~Lombardi, J.~Saragih, T.~Simon, and Y.~Sheikh, ``Deep appearance models for
  face rendering,'' \emph{ACM Trans. Graph.}, vol.~37, no.~4, pp. 68:1--68:13,
  Jul. 2018.

\bibitem{PGZED19}
\BIBentryALTinterwordspacing
J.~Philip, M.~Gharbi, T.~Zhou, A.~Efros, and G.~Drettakis, ``Multi-view
  relighting using a geometry-aware network,'' \emph{ACM Transactions on
  Graphics (SIGGRAPH Conference Proceedings)}, vol.~38, no.~4, July 2019.
  [Online]. Available: \url{http://www-sop.inria.fr/reves/Basilic/2019/PGZED19}
\BIBentrySTDinterwordspacing

\bibitem{nerf}
B.~Mildenhall, P.~P. Srinivasan, M.~Tancik, J.~T. Barron, R.~Ramamoorthi, and
  R.~Ng, ``{NeRF}: Representing scenes as neural radiance fields for view
  synthesis,'' in \emph{The European Conference on Computer Vision (ECCV)},
  2020.

\bibitem{nerv}
P.~P. Srinivasan, B.~Deng, X.~Zhang, M.~Tancik, B.~Mildenhall, and J.~T.
  Barron, ``Nerv: Neural reflectance and visibility fields for relighting and
  view synthesis,'' 2020.

\bibitem{boss2021nerd}
M.~Boss, R.~Braun, V.~Jampani, J.~T. Barron, C.~Liu, and H.~P. Lensch, ``Nerd:
  Neural reflectance decomposition from image collections,'' in \emph{IEEE
  International Conference on Computer Vision (ICCV)}, 2021.

\bibitem{kuang2021neroic}
Z.~Kuang, K.~Olszewski, M.~Chai, Z.~Huang, P.~Achlioptas, and S.~Tulyakov,
  ``{NeROIC}: Neural object capture and rendering from online image
  collections,'' \emph{Computing Research Repository (CoRR)}, vol.
  abs/2201.02533, 2022.

\bibitem{https://doi.org/10.48550/arxiv.2203.15946}
\BIBentryALTinterwordspacing
K.~Tiwary, T.~Klinghoffer, and R.~Raskar, ``Towards learning neural
  representations from shadows,'' 2022. [Online]. Available:
  \url{https://arxiv.org/abs/2203.15946}
\BIBentrySTDinterwordspacing

\bibitem{https://doi.org/10.48550/arxiv.2104.08418}
\BIBentryALTinterwordspacing
C.~Xie, K.~Park, R.~Martin-Brualla, and M.~Brown, ``Fig-nerf: Figure-ground
  neural radiance fields for 3d object category modelling,'' 2021. [Online].
  Available: \url{https://arxiv.org/abs/2104.08418}
\BIBentrySTDinterwordspacing

\bibitem{https://doi.org/10.48550/arxiv.2104.01148}
\BIBentryALTinterwordspacing
K.~Stelzner, K.~Kersting, and A.~R. Kosiorek, ``Decomposing 3d scenes into
  objects via unsupervised volume segmentation,'' 2021. [Online]. Available:
  \url{https://arxiv.org/abs/2104.01148}
\BIBentrySTDinterwordspacing

\bibitem{klinghoffer22physically}
T.~Klinghoffer, K.~Tiwary, A.~Balata, V.~Sharma, and R.~Raskar, ``Physically
  disentangled representations,'' \emph{arXiv preprint arXiv:2204.05281}, 2022.

\bibitem{chakrabarti2016learning}
A.~Chakrabarti, ``Learning sensor multiplexing design through
  back-propagation,'' \emph{Advances in Neural Information Processing Systems},
  vol.~29, 2016.

\bibitem{Sitzmann:2018:EndToEndCam}
V.~Sitzmann, S.~Diamond, Y.~Peng, X.~Dun, S.~Boyd, W.~Heidrich, F.~Heide, and
  G.~Wetzstein, ``End-to-end optimization of optics and image processing for
  achromatic extended depth of field and super-resolution imaging,'' \emph{ACM
  Trans. Graph. (SIGGRAPH)}, 2018.

\bibitem{metzler2020deep}
C.~A. Metzler, H.~Ikoma, Y.~Peng, and G.~Wetzstein, ``Deep optics for
  single-shot high-dynamic-range imaging,'' in \emph{Proceedings of the
  IEEE/CVF Conference on Computer Vision and Pattern Recognition}, 2020, pp.
  1375--1385.

\bibitem{chang2019deep}
J.~Chang and G.~Wetzstein, ``Deep optics for monocular depth estimation and 3d
  object detection,'' in \emph{Proceedings of the IEEE/CVF International
  Conference on Computer Vision}, 2019, pp. 10\,193--10\,202.

\bibitem{haim2018depth}
H.~Haim, S.~Elmalem, R.~Giryes, A.~M. Bronstein, and E.~Marom, ``Depth
  estimation from a single image using deep learned phase coded mask,''
  \emph{IEEE Transactions on Computational Imaging}, vol.~4, no.~3, pp.
  298--310, 2018.

\bibitem{he2018learning}
L.~He, G.~Wang, and Z.~Hu, ``Learning depth from single images with deep neural
  network embedding focal length,'' \emph{IEEE Transactions on Image
  Processing}, vol.~27, no.~9, pp. 4676--4689, 2018.

\bibitem{chang2018hybrid}
J.~Chang, V.~Sitzmann, X.~Dun, W.~Heidrich, and G.~Wetzstein, ``Hybrid
  optical-electronic convolutional neural networks with optimized diffractive
  optics for image classification,'' \emph{Scientific reports}, vol.~8, no.~1,
  pp. 1--10, 2018.

\bibitem{robidoux2021end}
N.~Robidoux, L.~E.~G. Capel, D.-e. Seo, A.~Sharma, F.~Ariza, and F.~Heide,
  ``End-to-end high dynamic range camera pipeline optimization,'' in
  \emph{Proceedings of the IEEE/CVF Conference on Computer Vision and Pattern
  Recognition}, 2021, pp. 6297--6307.

\bibitem{del2020learned}
P.~Del~Hougne, M.~F. Imani, A.~V. Diebold, R.~Horstmeyer, and D.~R. Smith,
  ``Learned integrated sensing pipeline: Reconfigurable metasurface
  transceivers as trainable physical layer in an artificial neural network,''
  \emph{Advanced Science}, vol.~7, no.~3, p. 1901913, 2020.

\bibitem{marco2017deeptof}
J.~Marco, Q.~Hernandez, A.~Munoz, Y.~Dong, A.~Jarabo, M.~H. Kim, X.~Tong, and
  D.~Gutierrez, ``Deeptof: off-the-shelf real-time correction of multipath
  interference in time-of-flight imaging,'' \emph{ACM Transactions on Graphics
  (ToG)}, vol.~36, no.~6, pp. 1--12, 2017.

\bibitem{su2018deep}
S.~Su, F.~Heide, G.~Wetzstein, and W.~Heidrich, ``Deep end-to-end
  time-of-flight imaging,'' in \emph{Proceedings of the IEEE Conference on
  Computer Vision and Pattern Recognition}, 2018, pp. 6383--6392.

\bibitem{nehme2019dense}
E.~Nehme, D.~Freedman, R.~Gordon, B.~Ferdman, T.~Michaeli, and Y.~Shechtman,
  ``Dense three dimensional localization microscopy by deep learning,''
  \emph{arXiv preprint arXiv:1906.09957}, 2019.

\bibitem{hershko2019multicolor}
E.~Hershko, L.~E. Weiss, T.~Michaeli, and Y.~Shechtman, ``Multicolor
  localization microscopy and point-spread-function engineering by deep
  learning,'' \emph{Optics express}, vol.~27, no.~5, pp. 6158--6183, 2019.

\bibitem{horstmeyer2017convolutional}
R.~Horstmeyer, R.~Y. Chen, B.~Kappes, and B.~Judkewitz, ``Convolutional neural
  networks that teach microscopes how to image,'' \emph{arXiv preprint
  arXiv:1709.07223}, 2017.

\bibitem{kellman2019data}
M.~Kellman, E.~Bostan, M.~Chen, and L.~Waller, ``Data-driven design for fourier
  ptychographic microscopy,'' in \emph{2019 IEEE International Conference on
  Computational Photography (ICCP)}.\hskip 1em plus 0.5em minus 0.4em\relax
  IEEE, 2019, pp. 1--8.

\bibitem{turpin2018light}
A.~Turpin, I.~Vishniakou, and J.~D. Seelig, ``Light scattering control with
  neural networks in transmission and reflection,'' \emph{Arxiv: 180505602
  [Cs]}, 2018.

\bibitem{ikoma2021depth}
H.~Ikoma, C.~M. Nguyen, C.~A. Metzler, Y.~Peng, and G.~Wetzstein, ``Depth from
  defocus with learned optics for imaging and occlusion-aware depth
  estimation,'' in \emph{2021 IEEE International Conference on Computational
  Photography (ICCP)}.\hskip 1em plus 0.5em minus 0.4em\relax IEEE, 2021, pp.
  1--12.

\bibitem{tseng2021differentiable}
E.~Tseng, A.~Mosleh, F.~Mannan, K.~St-Arnaud, A.~Sharma, Y.~Peng, A.~Braun,
  D.~Nowrouzezahrai, J.-F. Lalonde, and F.~Heide, ``Differentiable compound
  optics and processing pipeline optimization for end-to-end camera design,''
  \emph{ACM Transactions on Graphics (TOG)}, vol.~40, no.~2, pp. 1--19, 2021.

\bibitem{nguyen2022learning}
C.~M. Nguyen, J.~N. Martel, and G.~Wetzstein, ``Learning spatially varying
  pixel exposures for motion deblurring,'' \emph{arXiv}, 2022.

\bibitem{li2020end}
Y.~Li, M.~Qi, R.~Gulve, M.~Wei, R.~Genov, K.~N. Kutulakos, and W.~Heidrich,
  ``End-to-end video compressive sensing using anderson-accelerated unrolled
  networks,'' in \emph{2020 IEEE International Conference on Computational
  Photography (ICCP)}.\hskip 1em plus 0.5em minus 0.4em\relax IEEE, 2020, pp.
  1--12.

\bibitem{ramanath2005color}
R.~Ramanath, W.~E. Snyder, Y.~Yoo, and M.~S. Drew, ``Color image processing
  pipeline,'' \emph{IEEE Signal Processing Magazine}, vol.~22, no.~1, pp.
  34--43, 2005.

\bibitem{nishimura2018automatic}
J.~Nishimura, T.~Gerasimow, R.~Sushma, A.~Sutic, C.-T. Wu, and G.~Michael,
  ``Automatic isp image quality tuning using nonlinear optimization,'' in
  \emph{2018 25th IEEE International Conference on Image Processing
  (ICIP)}.\hskip 1em plus 0.5em minus 0.4em\relax IEEE, 2018, pp. 2471--2475.

\bibitem{tseng2019hyperparameter}
E.~Tseng, F.~Yu, Y.~Yang, F.~Mannan, K.~S. Arnaud, D.~Nowrouzezahrai, J.-F.
  Lalonde, and F.~Heide, ``Hyperparameter optimization in black-box image
  processing using differentiable proxies.'' \emph{ACM Trans. Graph.}, vol.~38,
  no.~4, pp. 27--1, 2019.

\bibitem{jastrebski2006improving}
G.~A. Jastrebski and D.~V. Arnold, ``Improving evolution strategies through
  active covariance matrix adaptation,'' in \emph{2006 IEEE international
  conference on evolutionary computation}.\hskip 1em plus 0.5em minus
  0.4em\relax IEEE, 2006, pp. 2814--2821.

\bibitem{heide2014flexisp}
F.~Heide, M.~Steinberger, Y.-T. Tsai, M.~Rouf, D.~Pajak, D.~Reddy, O.~Gallo,
  J.~Liu, W.~Heidrich, K.~Egiazarian \emph{et~al.}, ``Flexisp: A flexible
  camera image processing framework,'' \emph{ACM Transactions on Graphics
  (ToG)}, vol.~33, no.~6, pp. 1--13, 2014.

\bibitem{jiang2017learning}
H.~Jiang, Q.~Tian, J.~Farrell, and B.~A. Wandell, ``Learning the image
  processing pipeline,'' \emph{IEEE Transactions on Image Processing}, vol.~26,
  no.~10, pp. 5032--5042, 2017.

\bibitem{chen2018learning}
C.~Chen, Q.~Chen, J.~Xu, and V.~Koltun, ``Learning to see in the dark,'' in
  \emph{Proceedings of the IEEE conference on computer vision and pattern
  recognition}, 2018, pp. 3291--3300.

\bibitem{diamond2021dirty}
S.~Diamond, V.~Sitzmann, F.~Julca-Aguilar, S.~Boyd, G.~Wetzstein, and F.~Heide,
  ``Dirty pixels: Towards end-to-end image processing and perception,''
  \emph{ACM Transactions on Graphics (TOG)}, vol.~40, no.~3, pp. 1--15, 2021.

\bibitem{farhat1985optical}
N.~H. Farhat, D.~Psaltis, A.~Prata, and E.~Paek, ``Optical implementation of
  the hopfield model,'' \emph{Applied optics}, vol.~24, no.~10, pp. 1469--1475,
  1985.

\bibitem{sui2020review}
X.~Sui, Q.~Wu, J.~Liu, Q.~Chen, and G.~Gu, ``A review of optical neural
  networks,'' \emph{IEEE Access}, vol.~8, pp. 70\,773--70\,783, 2020.

\bibitem{zhang2020image}
Y.~Zhang, W.~Chen, H.~Ling, J.~Gao, Y.~Zhang, A.~Torralba, and S.~Fidler,
  ``Image gans meet differentiable rendering for inverse graphics and
  interpretable 3d neural rendering,'' \emph{arXiv preprint arXiv:2010.09125},
  2020.

\bibitem{parisi2019continual}
G.~I. Parisi, R.~Kemker, J.~L. Part, C.~Kanan, and S.~Wermter, ``Continual
  lifelong learning with neural networks: A review,'' \emph{Neural Networks},
  vol. 113, pp. 54--71, 2019.

\end{thebibliography}

\ifpeerreview \else
%%%% For the camera ready version, please fill out this
%%%% biography. Your camera ready should be within a 12 page limit
%%%% including acknowledgments, references and biography.

% \includegraphics command within an optional argument. (You could
% create your own custom macro containing the \includegraphics command
% to make things simpler here.)
% \begin{IEEEbiography}[{\includegraphics[width=1in,height=1.25in,clip,keepaspectratio]{mshell}}]{Michael Shell}

% \end{IEEEbiography}

%\newpage


%\vfill

\fi

\begin{IEEEbiography}[{\includegraphics[width=1in,height=1.25in,clip,keepaspectratio]{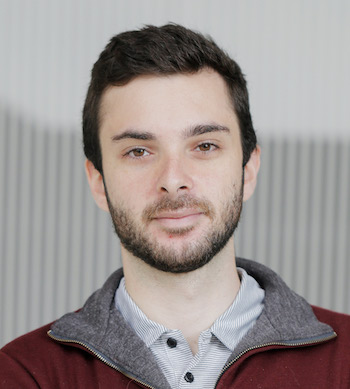}}]{Tzofi Klinghoffer} is a graduate student in the Camera Culture group at MIT Media Lab. He received his B.S. degree in computer science from The University of Alabama in 2018. He worked as an Associate Technical Staff member at MIT Lincoln Laboratory from 2018 to 2020 and as a software engineer on Alexa AI at Amazon from 2020 to 2021. His research interests include computer vision, imaging, and machine learning.
\end{IEEEbiography}

\begin{IEEEbiography}[{\includegraphics[width=1in,height=1.25in,clip,keepaspectratio]{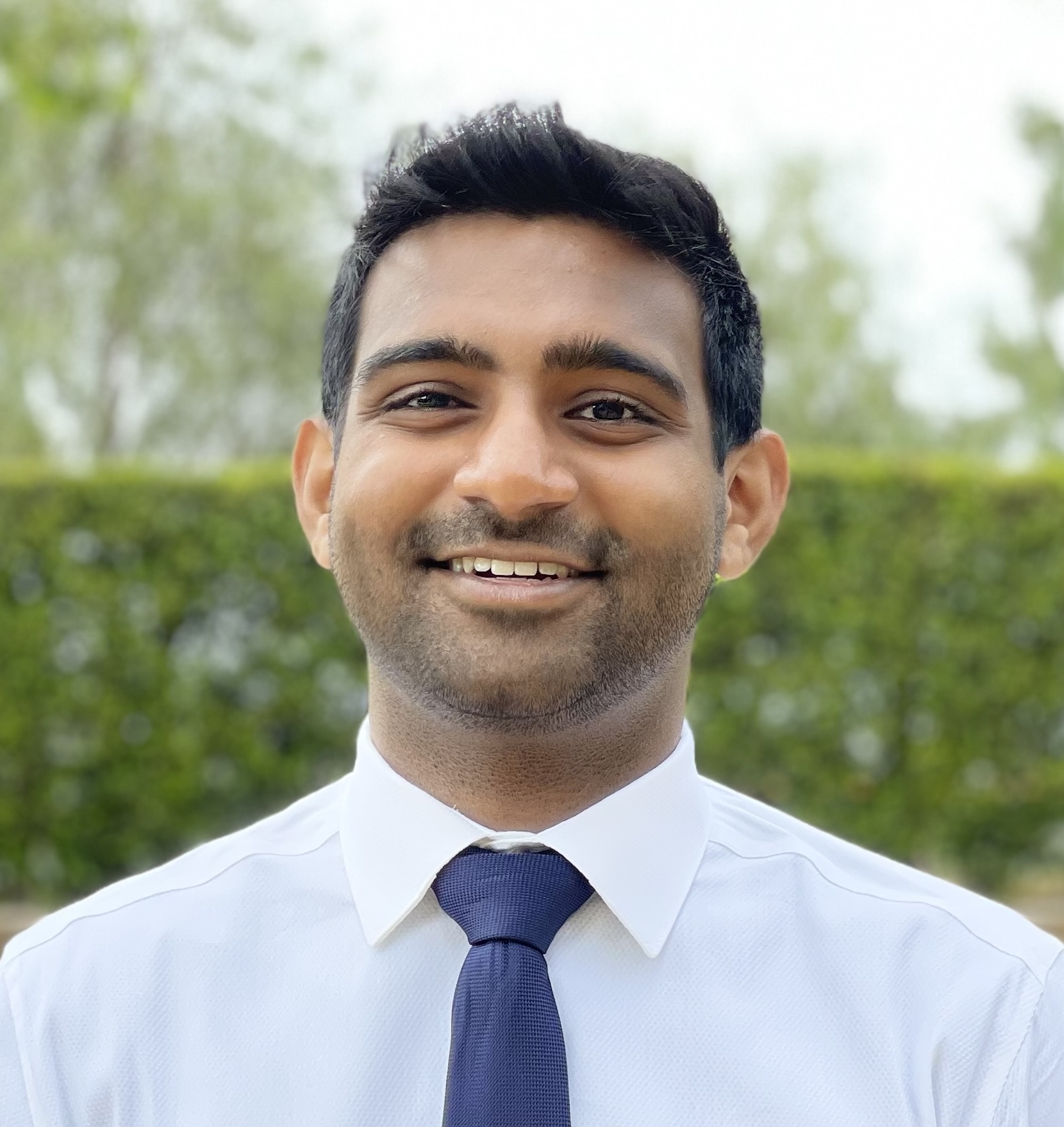}}]{Siddharth Somasundaram} is a graduate student in the Camera Culture group at MIT Media Lab, where he worked as research staff from 2021 to 2022. He received his B.S. degree in electrical engineering from University of California, Los Angeles in 2021. His past work spanned topics in photonics, optics, and computational imaging. His current research interests include  time-of-flight and non-line-of-sight imaging. 
\end{IEEEbiography}

\begin{IEEEbiography}[{\includegraphics[width=1in,height=1.25in,clip,keepaspectratio]{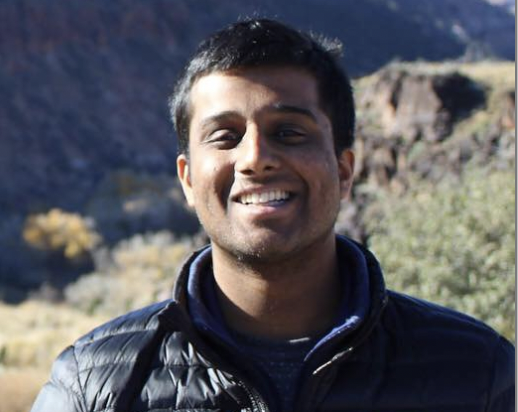}}]{Kushagra Tiwary} is a graduate student in the Camera Culture group at the MIT Media Lab. He received his B.S. degree from University of Illinois Urbana-Champaign in Electrical and Computer Engineering in 2019. Previously, he worked at Optimus Ride where he led the design and deployment of a fleet-wide MultiTasking Computer Vision System for next generation autonomous vehicles. His research interests are in imaging  perception and to build algorithms that exploit hidden cues present around us. 
\end{IEEEbiography}

\begin{IEEEbiography}[{\includegraphics[width=1in,height=1.25in,clip,keepaspectratio]{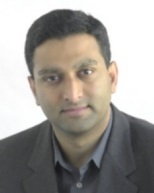}}]{Ramesh Raskar} is an Associate Professor at MIT Media Lab where he directs the Camera Culture research group. His focus is on AI and Imaging for health and sustainability. They span research in physical (e.g., sensors, health-tech), digital (e.g., automated and privacy-aware machine learning) and global (e.g., geomaps, autonomous mobility) domains. He received the Lemelson Award (2016), ACM SIGGRAPH Achievement Award (2017), DARPA Young Faculty Award (2009), Alfred P. Sloan Research Fellowship (2009), TR100 Award from MIT Technology Review (2004) and Global Indus Technovator Award (2003). He has worked on special research projects at Google [X] and Facebook and co-founded/advised several companies.
\end{IEEEbiography}

\end{document}